\documentclass{article}

\usepackage[preprint]{neurips_2025}

\usepackage[toc,page,header]{appendix}
\usepackage{minitoc}

\usepackage[utf8]{inputenc} 
\usepackage[T1]{fontenc}    
\usepackage{hyperref}       
\usepackage{url}            
\usepackage{booktabs}       
\usepackage{amsfonts}       
\usepackage{nicefrac}       
\usepackage{microtype}      
\usepackage{xcolor}         

\usepackage{tikz}
\usepackage{pgfplots}
\usepackage{xspace}
\usepackage{amsmath}
\usepackage{colortbl}
\usepackage{enumitem}
\usepackage[normalem]{ulem}
\usepackage{wrapfig}
\usepackage{caption} 
\usepackage{graphicx}
\usepackage{multirow}

\title{Boosting Generative Image Modeling via\\Joint Image-Feature Synthesis}

\author{%
  Theodoros Kouzelis\\
  Archimedes, Athena RC\\
  National Technical University of Athens\\
  \And
    Efstathios Karypidis\\
  Archimedes, Athena RC\\
  National Technical University of Athens\\
  \And
  Ioannis Kakogeorgiou\\
  Archimedes, Athena RC\\
  IIT, NCSR "Demokritos"\\
  \And
  Spyros Gidaris\\
  valeo.ai\\
  \And
  Nikos Komodakis\\
  Archimedes, Athena RC\\
  University of Crete\\
  IACM-Forth \\
}

\usetikzlibrary{arrows.meta,shapes,calc,matrix,fit,positioning,backgrounds,decorations.markings}
\pgfplotsset{compat=1.9}
\usepackage{xstring}

\usepgfplotslibrary{external}

\IfBeginWith*{\jobname}{fig/extern/}{\finalcopy}{}

\tikzstyle{every picture}+=[
	remember picture,
	every text node part/.style={align=center},
	every matrix/.append style={ampersand replacement=\&},
]
\tikzstyle{tight} = [inner sep=0pt,outer sep=0pt]
\tikzstyle{node}  = [draw,circle,tight,minimum size=12pt,anchor=center]
\tikzstyle{op}    = [draw,circle,tight]
\tikzstyle{dot}   = [fill,draw,circle,inner sep=1pt,outer sep=0]
\tikzstyle{pt}    = [fill,draw,circle,inner sep=1.5pt,outer sep=.2pt]
\tikzstyle{box}   = [draw,rectangle,inner sep=3pt]
\tikzstyle{high}  = [black!60]
\tikzstyle{group} = [high,box,opacity=.5]
\tikzstyle{dim1}  = [fill opacity=.3,text opacity=1]
\tikzstyle{dim2}  = [fill opacity=.5,text opacity=1]
\tikzstyle{dim3}  = [fill opacity=.7,text opacity=1]
\tikzstyle{rectc} = [tight,transform shape]
\tikzstyle{rect}  = [rectc,anchor=south west]

\tikzset{every mark/.append style={solid}}
\pgfplotsset{
	grid=both, width=\columnwidth, try min ticks=5,
	every axis/.append style={font=\small},
	every axis plot/.append style={thick,mark=none,mark size=1.8,tension=0.18},
	legend cell align=left, legend style={fill opacity=0.8},
	xticklabel={\pgfmathprintnumber[assume math mode=true]{\tick}},
	yticklabel={\pgfmathprintnumber[assume math mode=true]{\tick}},
	nodes near coords math/.style={
		nodes near coords={\pgfmathprintnumber[assume math mode=true]{\pgfplotspointmeta}},
	},
}

\pgfplotsset{
	dash/.style={mark=o,dashed,opacity=0.6},
	dott/.style={mark=o,dotted,opacity=0.6},
	nolim/.style={enlargelimits=false},
	plain/.style={every axis plot/.append style={},nolim,grid=none},
}

\pgfdeclarelayer{bg4}
\pgfdeclarelayer{bg3}
\pgfdeclarelayer{bg2}
\pgfdeclarelayer{bg1}
\pgfdeclarelayer{fg1}
\pgfdeclarelayer{fg2}
\pgfdeclarelayer{fg3}
\pgfdeclarelayer{fg4}
\pgfsetlayers{bg4,bg3,bg2,bg1,main,fg1,fg2,fg3,fg4}

\pgfkeys{/tikz/.cd, aspect/.store in=\aspect, aspect=1}
\pgfkeys{/tikz/.cd, depth/.store in=\depth, depth=.5}
\pgfkeys{/tikz/.cd, stepx/.store in=\step, stepx=1}

\tikzstyle{geom} = [line join=bevel,aspect=1,depth=.5,z={(\depth*\aspect,\depth)}]
\tikzstyle{wire} = [geom,draw,thick]

\def\cx[#1,#2,#3]{#1}
\def\cy[#1,#2,#3]{#2}
\def\cz[#1,#2,#3]{#3}
\def\ex[#1,#2,#3]{#1,0,0}
\def\ey[#1,#2,#3]{0,#2,0}
\def\ez[#1,#2,#3]{0,0,#3}

\newcommand{\Th}[1]{\textsc{#1}}

\newcommand{\red}[1]{{\textcolor{red}{#1}}}

\newcommand{\citeme}[1]{\red{[XX]}}
\newcommand{\refme}[1]{\red{(XX)}}

\makeatletter
\newcommand*\bdot{\mathpalette\bdot@{.7}}
\newcommand*\bdot@[2]{\mathbin{\vcenter{\hbox{\scalebox{#2}{$\m@th#1\bullet$}}}}}
\makeatother

\makeatletter
\DeclareRobustCommand\onedot{\futurelet\@let@token\@onedot}
\def\@onedot{\ifx\@let@token.\else.\null\fi\xspace}

\makeatother

\newcommand{\ditbtwo}{\texttt{DiT-B/2}\xspace}

\newcommand{\our}{\texttt{ReDi}\xspace}
\newcommand{\ours}{\texttt{ReDi}\xspace}

\definecolor{ForestGreen}{RGB}{34,139,34}
\definecolor{Orange}{RGB}{1.0, 0.55, 0.0}
\definecolor{Purple}{RGB}{0.58, 0.44, 0.86}
\definecolor{softblue}{rgb}{0.2, 0.4, 0.8} 
\definecolor{lightblue}{rgb}{0.6, 0.75, 0.95}
\definecolor{darkblue}{rgb}{0.05, 0.2, 0.5}

\definecolor{coral}{RGB}{255, 127, 80} 
\definecolor{softcoral}{rgb}{1.00, 0.60, 0.50}  

\definecolor{teal}{rgb}{0.1, 0.6, 0.6} 
\definecolor{forestgreen}{rgb}{0.1, 0.6, 0.2} 

\definecolor{vibrantorange}{rgb}{1.0, 0.5, 0.0}

\definecolor{deepblue}{rgb}{0.0, 0.0, 0.8}

\definecolor{TableColor}{rgb}{0.58, 0.55, 0.51}

\definecolor{TableColor}{rgb}{0.92, 0.95, 0.99}  
\definecolor{TableColorGrey}{rgb}{0.88, 0.88, 0.88}  

\definecolor{DarkenedMagenta}{RGB}{225,0,100}

\newcommand{\Eq}[1]{\hyperref[#1]{Eq.~(\ref{#1})}}
\newcommand{\Equation}[1]{\hyperref[#1]{Equation~(\ref{#1})}}
\newcommand{\secref}[1]{\hyperref[#1]{Sec.~\ref{#1}}}

\begin{document}

\maketitle

\addtocontents{toc}{\protect\setcounter{tocdepth}{-1}}

\begin{figure}[h]
        \centering
        \hspace*{-0.035\textwidth}%
        \includegraphics[width=0.95\textwidth]{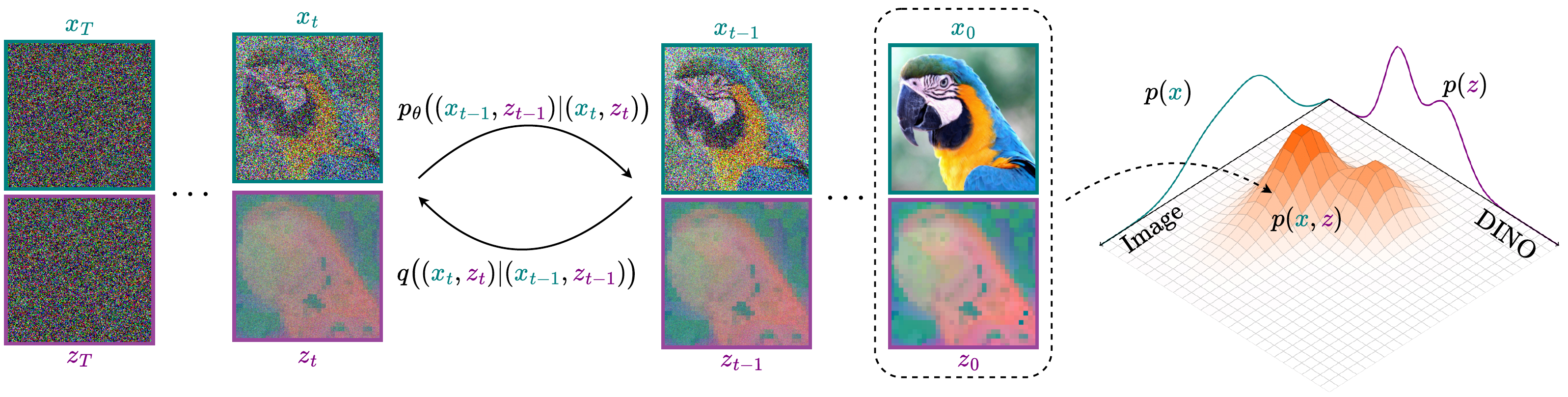} 
    \caption{\texttt{ReDi}: Our generative image modeling framework 
    bridges the gap between generative modeling and representation learning by leveraging a diffusion model that jointly captures low-level image details (via \texttt{VAE} latents) and high-level semantic features (via \texttt{DINOv2}). Trained to generate coherent image–feature pairs from pure noise, this 
    unified latent-semantic dual-space
    diffusion approach significantly boosts both generative quality and training convergence speed.}
    \label{fig:joint-diffusion}
\end{figure}
\begin{abstract}
Latent diffusion models (LDMs) dominate high-quality image generation, yet integrating representation learning with generative modeling remains a challenge. We introduce a novel generative image modeling framework that seamlessly bridges this gap by leveraging a diffusion model to jointly model low-level image latents (from a variational autoencoder) and high-level semantic features (from a pretrained self-supervised encoder like DINO). Our latent-semantic diffusion approach learns to generate coherent image–feature pairs from pure noise, significantly enhancing both generative quality and training efficiency, all while requiring only minimal modifications to standard Diffusion Transformer architectures. By eliminating the need for complex distillation objectives, our unified design simplifies training and unlocks a powerful new inference strategy: Representation Guidance, which leverages learned semantics to steer and refine image generation. Evaluated in both conditional and unconditional settings, our method delivers substantial improvements in image quality and training convergence speed, establishing a new direction for representation-aware generative modeling. Project page and code: \url{https://representationdiffusion.github.io/}
\end{abstract}

\section{Introduction}
\label{sec:intro}

\begin{figure}[t]
\vspace{0pt}%
\centering
\footnotesize

\begin{minipage}[t]{0.42\linewidth}
\centering
\begin{tikzpicture}[baseline=(current bounding box.north)]
    \hspace{5pt}
    \begin{axis}[
      scale only axis,
      width=0.95\linewidth,
      height=0.6\linewidth,
      xmin=4.2, xmax=10.3,
      ymin=1.3,  ymax=5.4,
      xtick={4.34194252, 4.88, 5.521234, 6.1234, 7.50159249, 8.45274191, 9.40389132, 10.17180649},
      xticklabels={100K,,,400K,1M,2M,4M,7M},
      ytick={1.60836753, 2.0, 3.03794033, 4.01668, 4.58920601, 4.99541967},
      yticklabels={3.3, 5,10,20,30,40},
      xlabel={\textbf{Training Iteration}},
      ylabel={\textbf{\Th{FID}}},
      y label style={
          at={(axis description cs:-0.1,0.6)},
          anchor=east
      },
      legend style={
          draw=black,
          line width=0.5pt,
          at={(0.51,0.992)},
          anchor=north west,
          font=\scriptsize
      },
      grid=major,
      major grid style={gray!20},
    ]

      \addplot[color=teal!30, mark=*, line width=1.2pt] 
        coordinates {
           (4.34194252, 5.25286188)
           (4.88, 4.88533619)
           (6.1234, 3.98093068)
           (10.17180649, 2.98029874)
        };
      \addlegendentry{\texttt{DiT-XL/2}}

      \addplot[color=teal, mark=*, line width=1.2pt] 
        coordinates {
           (4.34194252, 4.37624064)
           (4.88, 3.42996269)
           (5.521234, 3.05199042)
           (6.1234, 2.8412991)
        };
      \addlegendentry{\texttt{DiT-XL/2+}\our}

      \addplot[color=softblue!30, mark=*, line width=1.2pt]
        coordinates {
           (4.34194252, 5.0975381)
           (4.88, 4.6803358)
           (6.1234, 3.80371463)
           (10.17180649, 2.7748387)
        };
      \addlegendentry{\texttt{SiT-XL/2}}

      \addplot[color=softblue, mark=*, line width=1.2pt]
        coordinates {
           (4.34194252, 4.1954882)
           (4.88, 3.35302413)
           (6.1234, 2.63172666)
           (7.01215652, 2.21922335)
           (7.50159249, 2.03)
           (9.40389132, 1.60836753)
        };
      \addlegendentry{\texttt{SiT-XL/2+}\our}

      \draw[<-, dashed, black]
        (axis cs:5.9, 3.0) -- (axis cs:10.1, 2.995)
        node[midway, below=-0.7mm] {\(\times 23\) speed-up};


    \end{axis}
\end{tikzpicture}
\end{minipage}%
\hfill
\begin{minipage}[t]{0.42\linewidth}

\hspace{-50pt}
\centering
\begin{tikzpicture}[baseline=(current bounding box.north)]
    \hspace{5pt}
    \begin{axis}[
      scale only axis,
      width=0.95\linewidth,
      height=0.6\linewidth,
      xmin=4.2, xmax=9.8,
      ymin=1.3,  ymax=4.5,
      xtick={4.34194252, 4.88, 5.521234, 6.1234, 7.50159249, 8.45274191, 9.40389132, 10.17180649},
      xticklabels={100K,,,400K,1M,2M,4M},
      ytick={1.60836753, 2.0, 3.03794033, 4.01668, 4.58920601, 4.99541967},
      yticklabels={3.3, 5,10,20,30,40},
      xlabel={\textbf{Training Iteration}},
      ylabel={\textbf{\Th{FID}}},
      y label style={
          at={(axis description cs:-0.1,0.6)},
          anchor=east
      },
      legend style={
          draw=black,
          line width=0.5pt,
          at={(0.395,0.992)},
          anchor=north west,
          font=\scriptsize
      },
      grid=major,
      major grid style={gray!20},
    ]

      \addplot[color=lightblue, mark=*, line width=1.2pt]
        coordinates {
           (4.34194252, 3.9442527)
           (4.65, 3.17252048)
           (6.1234, 2.705095)
           (7.50159249, 2.40777273)
           (9.40389132, 2.29291083)
        };
      \addlegendentry{\texttt{SiT-XL/2+REPA}}

      \addplot[color=softblue, mark=*, line width=1.2pt]
        coordinates {
           (4.34194252, 4.1954882)
           (4.88, 3.35302413)
           (6.1234, 2.63172666)
           (7.50159249, 2.03)
           (9.40389132, 1.60836753)
        };
      \addlegendentry{\texttt{SiT-XL/2+}\our}

      \addplot[color=darkblue, mark=*, line width=1.2pt]
        coordinates {
           (4.34194252, 3.5054882)
           (4.88, 2.65302413)
           (6.1234, 2.1)
           (7.50159249, 1.7)
        };
      \addlegendentry{\texttt{SiT-XL/2+REPA+}\textbf{\texttt{ReDi}}}

      \draw[<-, dashed, softblue]
        (axis cs:7.29, 2.2) -- (axis cs:9.45, 2.2)
        node[pos=0.6, below=-0.12mm] {\(\times 6\) };

\draw[<-, dashed, darkblue]
  (axis cs:5.8, 2.39) -- (axis cs:9.45, 2.39)
  node[pos=-0.12, below=1.9mm] {\(\times 11 \)};

    \end{axis}
\end{tikzpicture}
\end{minipage}

\caption{
\textbf{Accelerated Training.} Generative performance curves on Imagenet $256 \times 256$ without Classifier-Free Guidance. \textbf{Left}:
 Our \our accelerates convergence of \texttt{DiT-XL/2} and \texttt{SiT-XL/2} by approximately $\times 23$. \textbf{Right:} \texttt{ReDi} converges $\times 6$ faster than \texttt{REPA}. When applied on top of \texttt{REPA} delivers a $\times 11$ speed-up.
}
\label{fig:convergence}
\end{figure}

Latent diffusion models (\texttt{LDMs})~\citep{rombach2022high} have emerged as a leading approach for high-quality image synthesis, achieving state-of-the-art results~\citep{rombach2022high, peebles2023scalable, ma2024sit}. These models operate in two stages: first, a variational autoencoder (\texttt{VAE}) compresses images into a compact latent representation \citep{rombach2022high}; second, a diffusion model learns the distribution of these latents, capturing their underlying structure.

Leveraging their intermediate features, pretrained \texttt{LDMs} have shown promise for various scene understanding tasks, including classification~\citep{mukhopadhyay2023diffusion}, pose estimation~\citep{gong2023diffpose}, and segmentation~\citep{li2023grounded,liu2023beyond,delatolas2025studyingimagediffusionfeatures}. However, their discriminative capabilities typically underperform specialized (self-supervised) representation learning approaches like masking-based~\citep{mae}, contrastive~\citep{simclr}, self-distillation~\citep{dino}, or vision-language contrastive~\citep{clip} methods. This limitation stems from the inherent tension in \texttt{LDM} training - the need to maintain precise low-level reconstruction while simultaneously developing semantically meaningful representations.

This observation raises a fundamental question: \textit{How can we leverage representation learning to enhance generative modeling?} Recent work by \cite{Yu2025repa} (\texttt{REPA}) demonstrates that improving the semantic quality of diffusion features through distillation of pretrained self-supervised representations leads to better generation quality and faster convergence. Their results establish a clear connection between representation learning and generative performance.

Motivated by these insights, we investigate whether a more effective approach to leveraging representation learning can further enhance image generation performance. In this work, we contend that the answer is \textit{yes}: rather than aligning diffusion features with external representations via distillation, we propose to \textit{jointly model both images (specifically their \texttt{VAE} latents) and their high-level semantic features} extracted from a pretrained vision encoder (e.g., \texttt{DINOv2}~\citep{oquab2024dinov}) within the same diffusion process. 
Formally, as shown in~\autoref{fig:joint-diffusion}, we define the forward diffusion process as $q(\mathbf{x}_t, \mathbf{z}_t | \mathbf{x}_{t-1}, \mathbf{z}_{t-1})$ for $t = 1, ..., T$, where $\mathbf{x}_0=\mathbf{x}$ and $\mathbf{z}_0=\mathbf{z}$ are the clean \texttt{VAE} latents and semantic features, respectively. The reverse process $p_{\theta}(\mathbf{x}_{t-1}, \mathbf{z}_{t-1}|\mathbf{x}_t, \mathbf{z}_t)$ learns to gradually denoise both modalities from Gaussian noise.

This joint modeling approach forces the diffusion model to explicitly learn the joint distribution of both precise low-level (\texttt{VAE}) and high-level semantic (\texttt{DINOv2}) features. We implement this approach, called \ours (\underline{Re}presentation \underline{Di}ffusion), within the \texttt{DiT} \citep{peebles2023scalable} and \texttt{SiT} \citep{ma2024sit} frameworks with minimal modifications to their transformer architecture: we apply standard diffusion noise to both representations, combine them into a single set of tokens, and train the standard diffusion transformer architecture to denoise both components simultaneously.

\newpage

Compared to \texttt{REPA}, our joint modeling approach offers three key advantages. First, the diffusion process explicitly models both low-level and semantic features, enabling direct integration of these complementary representations. Second, our method simplifies training by eliminating the need for additional distillation objectives. Finally, during inference, our unified approach enables \textit{Representation Guidance} - where the model uses its learned semantic understanding to iteratively refine generated images, improving quality in both conditional and unconditional generation.

Our contributions can be summarized as follows:
\begin{enumerate}[leftmargin=20pt, topsep=0pt, parsep=0pt]
	\item We propose \ours, a novel and effective method that jointly models image-compressed latents and semantically rich representations within the diffusion process, significantly improving image synthesis performance.  
	\item We provide a concrete implementation of our approach for both diffusion (\texttt{DiT}) and flow-matching (\texttt{SiT}) frameworks, leveraging \texttt{DINOv2}~\citep{oquab2024dinov} as the source of high-quality semantic representations.
    \item We also introduce \emph{Representation Guidance}, which leverages the model’s semantic predictions during inference to refine outputs, further enhancing image generation quality.
    \item We demonstrate that our approach boosts performance in both conditional and unconditional generation, while significantly accelerating convergence (see~\autoref{fig:convergence}). 
\end{enumerate}
\section{Related work}
\label{sec:related}

\paragraph{Representation Learning.}
Various approaches aim to learn meaningful representations for downstream tasks, with self-supervised learning emerging as one of the most promising directions. Early approaches employed pretext tasks such as predicting image patch permutations~\citep{jigsaw} or rotation angles~\citep{rotations}, while more recent methods utilize contrastive learning~\citep{simclr,infonce,pirl}, clustering-based objectives~\citep{swav,caron2018deep,caron2019unsupervised}, and self-distillation techniques~\citep{byol,simsiam,dino,gidaris2021obow}. The introduction of transformers enabled Masked Image Modeling (MIM), introduced by BEiT~\citep{beit} and evolved through SimMIM~\citep{simmim}, MAE~\cite{mae}, AttMask~\citep{kakogeorgiou2022attmask}, iBOT~\citep{zhou2021ibot}, and MOCA~\citep{gidaris2024moca}, with DINOv2~\citep{oquab2024dinov} achieving state-of-the-art performance through scaled models and datasets.  
Separately, contrastive vision-language pretraining, initiated by CLIP~\citep{clip}, established powerful joint image-text representations. Subsequent models like SigLIP~\cite{siglip} and SigLIPv2~\citep{siglipv2} refined this framework through enhanced training techniques, excelling in zero-shot settings and image retrieval~\citep{ilias2025}. Building on these advances, we leverage pretrained DINOv2 visual representations to enhance image generative modeling performance.

\paragraph{Diffusion Models and Representation Learning}
Due to the success of diffusion models, many recent works leverage representations learned
from pre-trained diffusion models for downstream tasks \citep{fuest2024diffusion}.  In particular, intermediate U-Net \citep{ronneberger2015u} features have been shown to capture rich semantic information, enabling tasks such as semantic segmentation \citep{baranchuk2022labelefficient, zhao2023unleashing}, semantic correspondence \citep{luo2023diffusion, zhang2023tale, hedlin2023unsupervised}, depth estimation \citep{zhao2023unleashing}, and image editing \citep{tumanyan2023plug}. Furthermore, diffusion models have been used for knowledge transfer by distilling learned representations through teacher-student frameworks \citep{li2023dreamteacher} or refining them via reinforcement learning \citep{yang2023diffusion}. Other works have shown that diffusion models learn strong discriminative features that can be leveraged for classification \citep{mukhopadhyay2023diffusion, xiang2023denoising}.
In a complementary direction, REPA \citep{Yu2025repa} recently demonstrated that aligning the internal representations of DiT \citep{peebles2023scalable} with a powerful pre-trained visual encoder during training significantly improves generative performance. Motivated by this observation, we propose to integrate images and semantic representations into a joint learning process.

\paragraph{Multi-modal Generative Modeling}

Unifying the generation across diverse modalities has recently attracted widespread interest. Notably, CoDi \citep{tang2023any} leverages a diffusion model that enables
generation across text, image, video, and audio in an aligned latent space. A joint representation for different modalities has been shown to have great scalability properties \citep{mizrahi20234m}.
For video generation, WVD \citep{zhang2024world} incorporates explicit 3D supervision by learning the joint distribution of RGB  and XYZ
frames. 
To capture richer spatial semantics, GEM \citep{hassan2024gem} generates paired
images and depth maps.
 MT-Diffusion \citep{chen2024diffusion} learns to incorporate various multi-modal data types with a multitask loss including CLIP \citep{radford2021learning} image representations. However, they do not quantitatively assess how this impacts the generative performance. VideoJam \citep{chefer2025videojam} models a joint image-motion representation that boosts temporal coherence 
 and introduces a theoretically motivated Classifier-Free Guidance (CFG) \cite{ho2022classifier} variant to condition on both motion and text.
 Inspired by this approach and building on the standard CFG framework, we propose Representation Guidance, incorporating the visual representations as an additional guidance signal during inference.


\begin{figure}[t]
        \centering
        \includegraphics[width=1.0\textwidth]{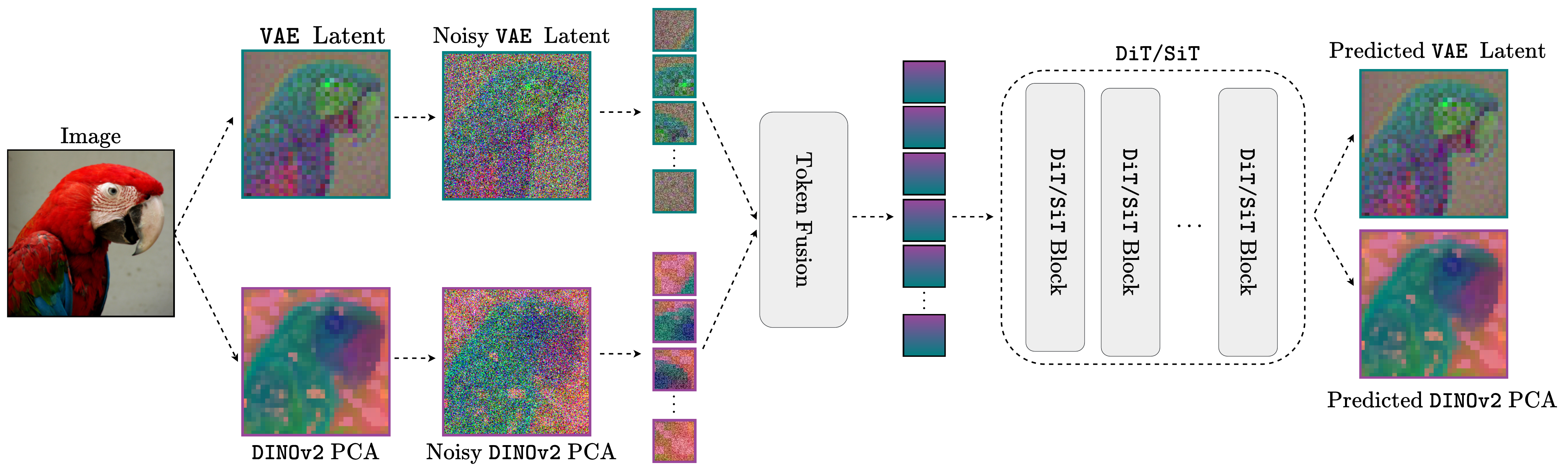} 

    \caption{Given an input image, the \texttt{VAE} latent and the principal components of \texttt{DINOv2} are extracted. 
    Both modalities are noised and fused into a \emph{joint token sequence}, given as input to \texttt{DiT} or \texttt{SiT}.}
    \label{fig:method}
    \vspace{-10pt}
\end{figure}
\section{Method}
\label{sec:d-dit}

\subsection{Preliminaries}
\label{sec:preliminaries}

\paragraph{Denoising Diffusion Probabilistic Models (DDPM)}
Diffusion models~\citep{ho2020denoising} generate data by gradually denoising a noisy input. The forward process corrupts an input $\mathbf{x}_0$ (e.g., an image or its VAE latent) over $T$ steps by adding Gaussian noise:
\begin{equation}  
\label{eq:forward_proc_prel}
    \mathbf{x}_t = \sqrt{\bar{\alpha}_t}\mathbf{x}_0 + \sqrt{1-\bar{\alpha}_t} \boldsymbol{\epsilon},
\end{equation}
where $\mathbf{x}_t$ is the noisy input at step $t$, $\bar{\alpha}_t$ are constants that define the noise schedule, and $\boldsymbol{\epsilon} \sim \mathcal{N}(\mathbf{0}, \mathbf{I})$ is the Gaussian noise term. Following \cite{ho2020denoising}, the reverse process learns to denoise $\mathbf{x}_t$ by predicting the added noise $\boldsymbol{\epsilon}$ using a network $\boldsymbol{\epsilon}_\theta(\cdot)$ with parameters $\theta$. The training objective is:
\begin{equation}  
    \mathcal{L}_{simple} = \mathbb{E}_{\mathbf{x}_0, \boldsymbol{\epsilon}, t} \Vert \boldsymbol{\epsilon}_\theta(\mathbf{x}_t, t) - \boldsymbol{\epsilon} \Vert^2.
\end{equation}
Although we also include the variational lower bound loss from \cite{nichol21a} to learn the variance of the reverse process, we omit it hereafter for brevity.

Unless otherwise specified, we focus on class-conditional image generation throughout this work. For notational simplicity, we omit explicit class conditioning variables from all mathematical formulations.

\paragraph{Diffusion Transformers (DiT)} 
The DiT \cite{peebles2023scalable} implements $\boldsymbol{\epsilon}_\theta$ using a Vision Transformer \cite{dosovitskiy2021image}. Given the ``patchified'' input $\mathbf{x}_t \in \mathbb{R}^{L \times C_x}$ ($L$ tokens of dimension $C_x$), the model first computes embeddings:
\begin{equation}  
    \mathbf{h}_t = \mathbf{x}_t \mathbf{W}_{emb}, \quad \mathbf{W}_{emb} \in \mathbb{R}^{C_x \times C_d}.
\end{equation}
The transformer processes $\mathbf{h}_t \in \mathbb{R}^{L \times C_d}$ to produce $\mathbf{o}_t \in \mathbb{R}^{L \times C_d}$. The final noise prediction is computed as:
\begin{equation}  
    \boldsymbol{\epsilon}_\theta(\mathbf{x}_t, t) = \mathbf{o}_t \mathbf{W}_{dec}, \quad \mathbf{W}_{dec} \in \mathbb{R}^{C_d \times C_x}.
\end{equation}

\subsection{Joint Image-Representation Generation}
\label{sec:joint-generation}

Our goal is to train a single model to jointly generate images and their semantic-aware visual representations by modeling their shared probability distribution. 
This approach captures the interdependent structures and features of both modalities. 
While we frame our approach using DDPM, it is also applicable to models trained with flow-matching objectives \cite{ma2024sit} (see \autoref{sec:sit_redi}).

A high-level overview of our method is depicted in \autoref{fig:method}.
Let $\text{I}$ denote a clean image, $\mathbf{x}_0 = \mathcal{E}_x(\text{I}) \in \mathbb{R}^{L \times C_x}$ its VAE tokens (produced by the VAE encoder $\mathcal{E}_x(\cdot)$), and $\mathbf{z}_0 = \mathcal{E}_z(\text{I}) \in \mathbb{R}^{L \times C_z}$ its patch-wise visual representation tokens (extracted by a pretrained encoder $\mathcal{E}_z(\cdot)$, e.g., DINOv2 \cite{oquab2024dinov})\footnote{For notational clarity, we incorporate the patchification step (typically with $2\times2$ patches in DiT architectures) into the encoder definitions $\mathcal{E}_x$ and $\mathcal{E}_z$.}.
To match the spatial resolution of $\mathbf{x}_0$, we assume $\mathcal{E}_z(\cdot)$ includes a bilinear resizing operation.  

During training, given $\mathbf{x}_0$ and $\mathbf{z}_0$, we define a joint forward diffusion processes:  
\begin{equation}  
\label{eq:forward_proc}  
    \mathbf{x}_t = \sqrt{\bar{\alpha}_t}\mathbf{x}_0 + \sqrt{1-\bar{\alpha}_t} \boldsymbol{\epsilon}_x, \quad  
    \mathbf{z}_t = \sqrt{\bar{\alpha}_t}\mathbf{z}_0 + \sqrt{1-\bar{\alpha}_t} \boldsymbol{\epsilon}_z, 
\end{equation}  
where $\bar{\alpha}_t$ controls the noise schedule and $\boldsymbol{\epsilon}_x \sim \mathcal{N}(\mathbf{0}, \mathbf{I})$, $\boldsymbol{\epsilon}_z \sim \mathcal{N}(\mathbf{0}, \mathbf{I})$ are Gaussian noise terms of dimensions $\mathbb{R}^{L \times C_x}$ and $\mathbb{R}^{L \times C_z}$, respectively.

The diffusion model $\boldsymbol{\epsilon}_\theta(\mathbf{x}_t, \mathbf{z}_t, t)$ takes as input $\mathbf{x}_t$ and $\mathbf{z}_t$, along with timestep $t$, and jointly predicts the noise for both inputs. Specifically, it produces two separate predictions: $\textcolor{black}{\boldsymbol{\epsilon}^x_\theta}(\mathbf{x}_t, \mathbf{z}_t, t)$ for the image latent noise $\boldsymbol{\epsilon}_x$,  and $\textcolor{black}{\boldsymbol{\epsilon}^z_\theta}(\mathbf{x}_t, \mathbf{z}_t, t)$ for the visual representation noise $\boldsymbol{\epsilon}_z$. The training objective combines both predictions:
\begin{equation}  
    \label{eq:joint_objective}  
    \mathcal{L}_{joint} = \underset{\mathbf{x}_0, \mathbf{z}_0, t} { \mathbb{E}} \Big [  
    \Vert \textcolor{black}{\boldsymbol{\epsilon}^x_\theta}(\mathbf{x}_t,\mathbf{z}_t, t) - \boldsymbol{\epsilon}_x \Vert^2  
    + \lambda_z \Vert \textcolor{black}{\boldsymbol{\epsilon}^z_\theta}(\mathbf{x}_t,\mathbf{z}_t, t) - \boldsymbol{\epsilon}_z \Vert^2 \Big],  
\end{equation}  
where $\lambda_z$ balances the denoising loss for $\mathbf{z_t}$. By default, we use $\lambda_z = 1$ in our experiments.

\begin{wrapfigure}{r}{0.5\textwidth}
  \vspace{-40pt}
  \centering
  \includegraphics[width=0.48\textwidth]{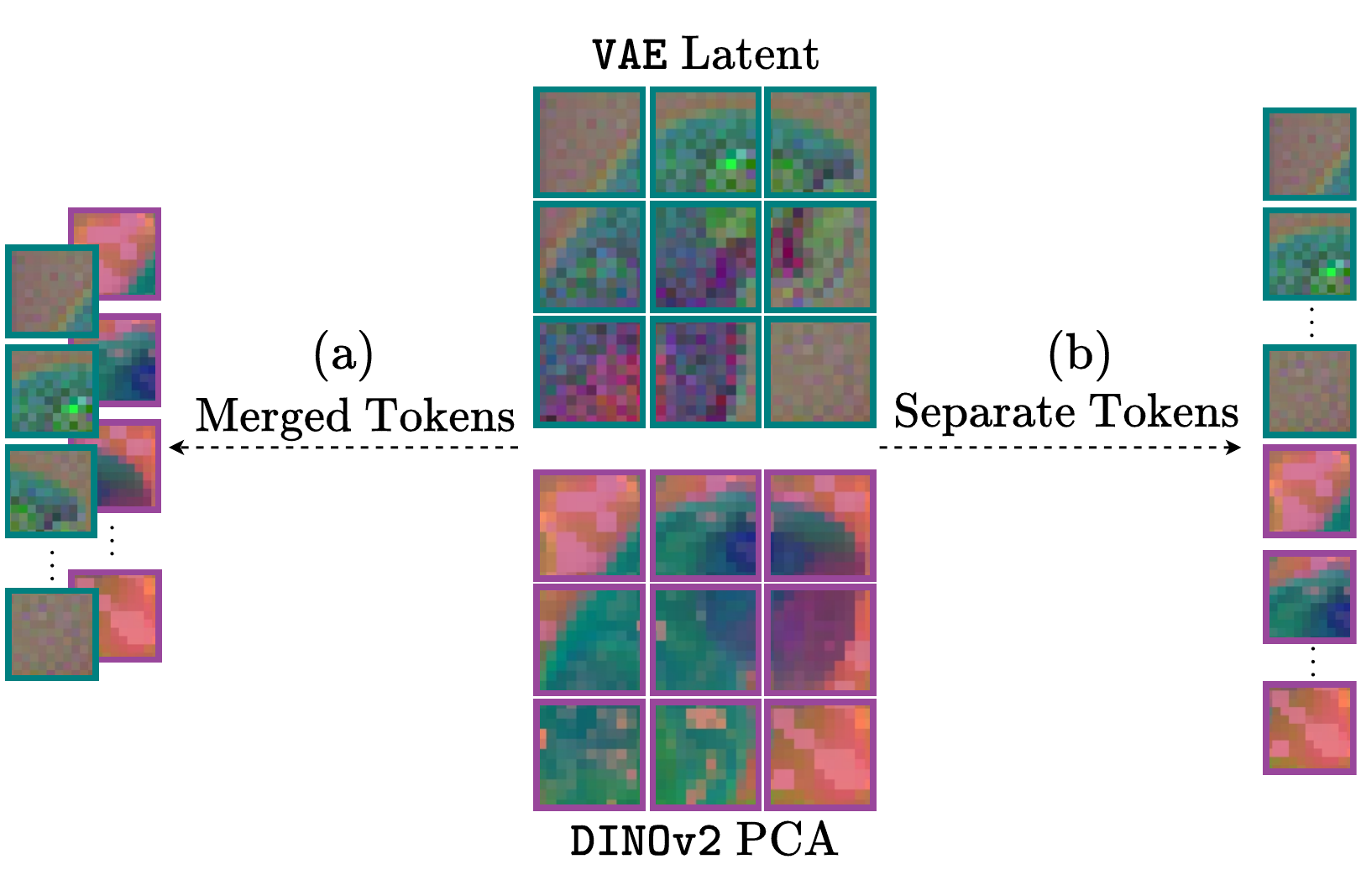}
  \caption{An illustration of our proposed token fusion approaches: (a) The tokens of the $\texttt{VAE}$ latents and the $\texttt{DINOv2}$ are merged channel-wise, (b) The tokens are concatenated along the sequence dimension.}
  \label{fig:agg}
\end{wrapfigure}

\subsection{Fusion of Image and Representation Tokens} 
\label{sec:agg}

We explore two approaches to combine and jointly process $\mathbf{x}_t$ and $\mathbf{z}_t$ in the diffusion transformer architecture: (1) merging tokens along the embedding dimension, and (2) maintaining separate tokens for each modality (see Fig.~\ref{fig:agg}). Both methods require only minimal modifications to the DiT architecture, specifically defining modality-specific embedding matrices $\mathbf{W}_{\text{emb}}^x \in \mathbb{R}^{C_x \times C_d}$ and $\mathbf{W}_{\text{emb}}^z \in \mathbb{R}^{C_z \times C_d}$, along with prediction heads $\mathbf{W}_{\text{dec}}^x \in \mathbb{R}^{C_d \times C_x}$ and $\mathbf{W}_{\text{dec}}^z \in \mathbb{R}^{C_d \times C_z}$ for $\mathbf{x}_t$ and $\mathbf{z}_t$ respectively.

\paragraph{Merged Tokens} 
The tokens are embedded separately and summed channel-wise:
\begin{equation}
\mathbf{h}_t = \mathbf{x}_t \mathbf{W}_{\text{emb}}^x + \mathbf{z}_t \mathbf{W}_{\text{emb}}^z \in \mathbb{R}^{L \times C_d}.  
\end{equation}
The transformer processes $\mathbf{h}_t$ to produce $\mathbf{o}_t$, with predictions:
\begin{equation}
\boldsymbol{\epsilon}^x_\theta = \mathbf{o}_t  \mathbf{W}_{\text{dec}}^x, \quad  
\boldsymbol{\epsilon}^z_\theta = \mathbf{o}_t \mathbf{W}_{\text{dec}}^z.  
\end{equation}
This approach enables early fusion while maintaining computational efficiency, as the token count remains unchanged.

\paragraph{Separate Tokens} 
Tokens are embedded separately and concatenated along the sequence dimension:
\begin{equation}
\mathbf{h}_t = [\mathbf{x}_t \mathbf{W}_{\text{emb}}^x \,, \, \mathbf{z}_t \mathbf{W}_{\text{emb}}^z] \in \mathbb{R}^{2L \times C_d},  
\end{equation}
where $[\cdot \,, \, \cdot]$ denotes sequence-wise concatenation. The transformer outputs separate representations $\mathbf{o}_t = [\mathbf{o}_t^x \,, \, \mathbf{o}_t^z]$, with predictions:
\begin{equation}
\boldsymbol{\epsilon}^x_\theta = \mathbf{o}_t^x \mathbf{W}_{\text{dec}}^x, \quad  
\boldsymbol{\epsilon}^z_\theta = \mathbf{o}_t^z \mathbf{W}_{\text{dec}}^z.  
\end{equation}
This method provides greater expressive power by preserving modality-specific information throughout processing, at the cost of increased computation due to increased token count.

Unless stated otherwise, we use the merged tokens approach for computational efficiency.

\subsection{Dimensionality-Reduced Visual Representation} \label{sec:pca_reduced}

In practice, the channel dimension of visual representations ($C_z$) significantly exceeds that of image latents ($C_x$), i.e., $C_z \gg C_x$. We empirically observe that this imbalance degrades performance, as the model disproportionately allocates capacity to visual representations at the expense of image latents.  

To address this, we apply Principal Component Analysis (PCA) to reduce the dimensionality of $\mathbf{z}_0$ from $C_z$ to $C'_z$ (where $C'_z \ll C_z$), preserving essential information while simplifying the prediction task. The PCA projection matrix is precomputed using visual representations sampled from the training set. All visual representations in Sections \ref{sec:joint-generation} and \ref{sec:agg} refer to these PCA-reduced versions.

\subsection{Representation Guidance} \label{sec:representation_guidance}

To ensure the generated images remain strongly influenced by the visual representations during inference, we introduce Representation Guidance. This technique during inference modifies the posterior distribution to: $\hat{p}_\theta(\mathbf{x}_t, \mathbf{z}_t) \propto p_\theta(\mathbf{x}_t) p( \mathbf{z}_t \vert \mathbf{x}_t)^{w_r}$, where $w_r$ controls how strongly samples are pushed toward higher likelihoods of the conditional distribution $p_\theta(\mathbf{z}_t | \mathbf{x}_t)$. Taking the log derivative yields the guided score function:
\begin{align}
    \nabla_{\!\mathbf{x}_t} \text{log} \; \hat{p}_\theta(\mathbf{x}_t, \mathbf{z}_t)=&
        \nabla_{\!\mathbf{x}_t} \text{log} \;p_\theta(\mathbf{x}_t)+
        w_r\big(
    \nabla_{\!\mathbf{x}_t} \text{log} \;p_\theta(\mathbf{z}_t \vert \mathbf{x}_t)
        \big) \\
        =& \nabla_{\!\mathbf{x}_t} \text{log} \;p_\theta(\mathbf{x}_t)+
        w_r\big(
    \nabla_{\!\mathbf{x}_t} \text{log} \;p_\theta(\mathbf{x}_t, \mathbf{z}_t)-
    \nabla_{\!\mathbf{x}_t} \text{log} \;p_\theta(\mathbf{x}_t)
        \big).
\end{align}

By recalling the equivalence of denoisers and scores \citep{vincent2011}, we implement this representation-guided prediction $\boldsymbol{\hat{e}_\theta}(\mathbf{x}_t, \mathbf{z}_t, t)$ at each denoising step as follows:
\begin{equation}
\boldsymbol{\hat{\epsilon}}_\theta(\mathbf{x}_t, \mathbf{z}_t, t) = \boldsymbol{\epsilon}_\theta(\mathbf{x}_t, t) + w_r\left(\boldsymbol{\epsilon}_\theta(\mathbf{x}_t, \mathbf{z}_t, t) - \boldsymbol{\epsilon}_\theta(\mathbf{x}_t, t)\right).
\end{equation}
Following \citet{ho2022classifier}, we train both $\boldsymbol{e_\theta}(\mathbf{x}_t, \mathbf{z}_t, t)$ and $\boldsymbol{e_\theta}(\mathbf{x}_t, t)$ jointly. Specifically, during training, with probability $p_{drop}$, we zero out $\mathbf{z}_t$ (setting $\boldsymbol{\epsilon}_\theta(\mathbf{x}_t, t) = \boldsymbol{\epsilon}_\theta(\mathbf{x}_t, \mathbf{0}, t)$) and disable the visual representation denoising loss by setting $\lambda_z = 0$ in \autoref{eq:joint_objective}.

\section{Experiments}
\label{sec:experiments}

\subsection{Setup}

\paragraph{Implementation details.}  
We follow the standard training setup of \texttt{DiT} \citep{peebles2023scalable} and \texttt{SiT} \citep{ma2024sit}, training on ImageNet at $256\times256$ resolution with a batch size of 256. Following ADM's preprocessing pipeline \citep{dhariwal2021adm}, we center-crop and resize all images to $256\times256$. Our experiments utilize transformer architectures \texttt{B/2}, \texttt{L/2}, and \texttt{XL/2} all using a $2\times2$ patch size. 
For unconditional generation, we simply set the number of classes to 1, maintaining the original architecture.
Images are encoded into VAE latent representations using \texttt{SD-VAE-FT-EMA} \citep{rombach2022high} that produces outputs with $\times$8 spatial downsampling factor and 4 output channels. For $256\times256$ images, this results in $32\times32\times4$ latent features. Through patchification with $2\times2$ patches, the VAE encoder $\mathcal{E}_x(\cdot)$ yields $L=256$ tokens, each with $C_x=16$ channels (4 channels $\times$ 2$\times$2 patch size).
For semantic representation extraction, we employ \texttt{DINOv2-B} with registers \citep{darcet2023vision,oquab2024dinov}. The 768-dimensional embeddings are reduced to 8 dimensions via PCA (trained on 76,800 randomly sampled ImageNet images). After bilinear interpolation to match the VAE's $32\times32\times4$ spatial resolution and $2\times2$ patchification, the encoder $\mathcal{E}_z(\cdot)$ produces $L=256$ tokens with $C_z=32$ channels each (8 channels $\,\times\,$ 2$\times$2 patch size).

\paragraph{Sampling.}
For \texttt{DiT} models, we adopt DDPM sampling, while for \texttt{SiT} models, we employ the SDE Euler–Maruyama sampler. The number of sampling steps is fixed at $250$ across all experiments.
When using Classifier-Free Guidance (CFG) \citep{ho2022classifier}, we apply it only to the \texttt{VAE} channels, with a guidance scale of $w = 2.4$ (see \autoref{fig:cfg-ablation}).
For Representation Guidance, we set $p_{drop}=0.2$, the guidance scale to $w_r = 1.5$ for \texttt{B} models and $w_r = 1.1$ for \texttt{XL} models.

\paragraph{Evaluation.} To benchmark generative performance, we report  Frechet Inception Distance (FID)
\citep{heusel2017gans}, sFID \citep{nash2021generating}, Inception
Score (IS) \citep{salimans2016improved}, Precision (Pre.) and
Recall (Rec.) \citep{kynkaanniemi2019improved} using $50$k
samples and the ADM’s TensorFlow evaluation suite \citep{dhariwal2021adm}.




\begin{minipage}[t]{0.41\textwidth}

\captionof{table}{\textbf{\Th{FID} Comparisons.} \Th{FID} scores on ImageNet $256\times256$ without Classifier-Free Guidance for \texttt{DiT} and \texttt{SiT} models of various sizes with \texttt{REPA} and \texttt{ReDi} (ours).}

\vspace{-4pt}

    \centering
    \footnotesize
    \setlength{\tabcolsep}{2.5pt} 
\begin{tabular}{lccc}
\toprule
\Th{Model} & \Th{\#Params} & \Th{Iter.} & \Th{FID$\downarrow$} \\ 
\midrule 


\texttt{DiT-L/2} & $458\text{M}$ & $400\text{K}$ & $23.2$ \\
w/ \texttt{REPA}  & $458\text{M}$ & $400\text{K}$ & $15.6$ \\
\rowcolor{TableColor} w/ \texttt{ReDi} (ours)  & $458\text{M}$ & $400\text{K}$ & $10.5$ \\

\midrule 
\texttt{SiT-L/2} & $458\text{M}$ & $400\text{K}$ & $18.5$ \\
w/ \texttt{REPA}  & $458\text{M}$ & $400\text{K}$ & $9.7$ \\
\rowcolor{TableColor} w/ \texttt{ReDi} (ours)  & $458\text{M}$ & $400\text{K}$ & $9.4$ \\

\midrule 

\texttt{DiT-XL/2} & $675\text{M}$ & $400\text{K}$ & $19.5$ \\
w/ \texttt{REPA}  & $675\text{M}$ & $400\text{K}$ & $12.3$ \\

\arrayrulecolor{black!30}\cmidrule(lr){1-4}

\texttt{DiT-XL/2} & $675\text{M}$ & $7\text{M}$ & $9.6$ \\
w/ \texttt{REPA}  & $675\text{M}$ & $850\text{K}$ & $9.6$ \\
\rowcolor{TableColor} w/ \texttt{ReDi} (ours)  & $675\text{M}$ & $400\text{K}$ & $8.7$ \\

\arrayrulecolor{black}\midrule 

\texttt{SiT-XL/2} & $675\text{M}$ & $400\text{K}$ & $17.2$ \\
w/ \texttt{REPA}  & $675\text{M}$ & $400\text{K}$ & $7.9$ \\
\rowcolor{TableColor} w/ \texttt{ReDi} (ours)  & $675\text{M}$ & $400\text{K}$ & $7.5$ \\ 

\arrayrulecolor{black!30}\cmidrule(lr){1-4}

\texttt{SiT-XL/2} & $675\text{M}$ & $7\text{M}$ & $8.3$ \\

w/ \texttt{REPA}  & $675\text{M}$ & $4\text{M}$ & $5.9$ \\
\rowcolor{TableColor} w/ \texttt{ReDi} (ours)  & $675\text{M}$ & $700\text{K}$ & $5.6$ \\

\rowcolor{TableColor} w/ \texttt{ReDi} (ours)  & $675\text{M}$ & $4\text{M}$ & $3.3$ \\

\arrayrulecolor{black}\bottomrule

\end{tabular}

\label{tab:fid_comparison}
\end{minipage}
\hfill
\begin{minipage}[t]{0.55\textwidth}
\footnotesize


\captionof{table}{\textbf{Comparison with State-of-the-art.} Quantitative evaluation on ImageNet $256 \times 256$ with Classifier-Free Guidance. Both \texttt{REPA} and \texttt{ReDi} (ours) employ \texttt{SiT-XL/2} as the base model.}
\centering

\vspace{-4.pt}

\setlength{\tabcolsep}{2.5pt} 

\begin{tabular}{l c c c c c c}
\toprule
\Th{Model} & \Th{Epochs} & \Th{FID$\downarrow$} & \Th{sFID$\downarrow$} & \Th{IS$\uparrow$} & \Th{Pre.$\uparrow$} & \Th{Rec.$\uparrow$} \\
\arrayrulecolor{black}\midrule

\rowcolor{TableColorGrey}\multicolumn{7}{c}{\emph{Autoregressive Models}} \\

 \texttt{VAR}  &350 &  1.80 & - & 365.4  & 0.83 & 0.57 \\
 \texttt{MagViTv2}  &1080 &  1.78 & - & 319.4  & 0.83 & 0.57 \\
 \texttt{MAR}  &800 &  1.55 & - & 303.7  & 0.81 & 0.62 \\
 
\arrayrulecolor{black!40}\midrule

\rowcolor{TableColorGrey}\multicolumn{7}{c}{\emph{Latent Diffusion Models}} \\

 \texttt{LDM} & 200 & 3.60 & -  & 247.7 & 0.87 & 0.48 \\ 
 \texttt{U-ViT-H/2} & 240 & 2.29 & 5.68  & 263.9 & 0.82 & 0.57 \\ 
 \texttt{DiT-XL/2}   & 1400  &    2.27 & 4.60 & {278.2} & {0.83} & 0.57  \\
 \texttt{MaskDiT} & 1600 &  2.28 & 5.67 & 276.6 & 0.80 & 0.61 \\ 
 \texttt{SD-DiT} & 480 & 3.23 & -    & -     & -    & -     \\
 \texttt{SiT-XL/2}   & 1400 &     2.06 & {4.50} & 270.3 & 0.82 & 0.59 \\
  \texttt{FasterDiT}   & 400 &     2.03 & {4.63} & 264.0 & 0.81 & 0.60 \\
    \texttt{MDT}   & 1300 &     1.79 & {4.57} & 283.0 & 0.81 & 0.61 \\

 \arrayrulecolor{black!40}\midrule
 \rowcolor{TableColorGrey}\multicolumn{7}{c}{\emph{Leveraging Visual Representations}} \\

\texttt{REPA} & {800} & {1.80} & {{4.50}} & {{284.0}} & {0.81} & {0.61} \\
\rowcolor{TableColor}\texttt{ReDi} (ours) & 350 & 1.72 & {4.68} & {278.7}
& {0.77} & {0.63} \\
\rowcolor{TableColor}\texttt{ReDi} (ours) & 800 & 1.61 & {4.66} & {295.1}
& {0.78} & 0.64 \\

\arrayrulecolor{black}\bottomrule
\end{tabular}


\label{tab:sota_comparison}

\end{minipage}
\vspace{-9pt}


\subsection{Enhancing the performance of generative models}

\paragraph{DiT \& SiT.} 
To demonstrate the effectiveness of our approach, we present performance gains for various-sized \texttt{DiT} and \texttt{SiT} models in \autoref{tab:fid_comparison}. Our method, \texttt{ReDi}, consistently delivers substantial improvements across models of different scales. Notably, \texttt{DiT-XL/2} with \texttt{ReDi} achieves an FID of $8.7$ after just $400$k iterations, outperforming the baseline \texttt{DiT-XL/2} trained for $7$M steps. Similarly, \texttt{SiT-XL/2} with \texttt{ReDi} reaches an FID of $7.5$ at $400$k iterations, surpassing the converged \texttt{SiT-XL} at $7$M steps. 
Additionally, \autoref{tab:sota_comparison} reports results for \texttt{SiT-XL/2} with Classifier-Free Guidance (CFG) \cite{ho2022classifier}. Once again, \texttt{ReDi} yields significant improvements, achieving an FID of $1.72$ in just $350$ epochs, outperforming the baseline trained to convergence over $1400$ epochs.

\paragraph{Comparison with REPA.} 
We further compare our results with \texttt{REPA}, which also leverages \texttt{DINOv2} features to enhance generative performance. Our approach, \texttt{ReDi}, consistently achieves superior generative performance with both $\texttt{DiT}$ and \texttt{SiT} as the base models. As shown in \autoref{tab:fid_comparison}, \texttt{DiT-L/2} with \texttt{ReDi} achives an FID of 10.5 significantly outperforming \texttt{DiT-L/2} with \texttt{REPA}. Notably, it even surpasses \texttt{REPA} trained for the same number of iterations with the larger \texttt{DiT-XL/2}, which achieves a higher FID of $12.3$. Further for \texttt{SiT-XL} models, \texttt{ReDi} attains an FID of 5.6 in just $700$k iterations, while \texttt{REPA} requires $4\text{M}$ iterations to reach an FID of 5.9. These results highlight the effectiveness of our method in leveraging visual representations to significantly boost generative performance.

\paragraph{ReDi is complementary to REPA.} Interestingly, we observe that the joint modeling objective of our \texttt{ReDi} and the alignment objective of \texttt{REPA} are complementary. As presented in \autoref{tab:bench-redi-repa} \texttt{REPA} + \texttt{ReDi}  matches the FID of the fully-converged \texttt{REPA} after only $350$K iterations,  and at $1$M iterations reaches an FID of $3.6$. For the implementation details, see Appendix \ref{sec:further_impl}.

\paragraph{Accelerating convergence.} 
The aforementioned results indicate that \texttt{ReDi} significantly accelerates the convergence of latent diffusion models. As illustrated in \autoref{fig:convergence}, \texttt{ReDi} speeds up the convergence of \texttt{DiT-XL/2} and \texttt{SiT-XL/2} by approximately $\times23$, respectively. Even when compared with \texttt{REPA}, \texttt{ReDi} demonstrated a $\times6$ faster convergence. When \texttt{ReDi} is applied on top of \texttt{REPA}, the convergence is $\times 11$ faster.

\begin{table}[t]
\centering
\begin{minipage}[t]{0.48\textwidth}

\caption{\textbf{Unconditional Generation \Th{FID} Performance.} Results on  ImageNet $256\times256$.
For comparison, we include conditional generation results (shown in \textcolor{gray}{gray}). Models at 400K
steps. \texttt{RG} denotes using Representation Guidance. }
    \vspace{5pt}
    \centering
    \setlength{\tabcolsep}{4pt}
    \begin{tabular}{lccc}
    \toprule
    \Th{Model} & \Th{\#Params} & \Th{FID$\downarrow$} \\ 
    \midrule 
    \rowcolor{TableColorGrey} \ditbtwo (conditional) & $130\text{M}$ &  $43.5$ \\
    \ditbtwo & $130\text{M}$ &  $69.3$ \\
    w/ \texttt{ReDi} (ours)   & $130\text{M}$ & $51.7$ \\
    w/ \texttt{ReDi}+\texttt{RG} (ours)  & $130\text{M}$ & $47.3$ \\

    \midrule 
    \rowcolor{TableColorGrey} \texttt{DiT-XL/2} (conditional) & $675\text{M}$ &  $19.5$ \\
    \texttt{DiT-XL/2} & $675\text{M}$ &  $44.6$ \\
    w/ \texttt{ReDi} (ours)  & $675\text{M}$ &  $25.1$ \\
    w/ \texttt{ReDi}+\texttt{RG} (ours)  & $675\text{M}$ & $22.6$ \\
    \bottomrule
    \end{tabular}
    \vspace{3pt}

    \label{tab:bench-uncond}

\end{minipage}
\hfill
\begin{minipage}[t]{0.48\textwidth}

    \caption{\textbf{\Th{FID} with Representation Guidance.} \Th{FID} scores on ImageNet $256\times256$. \texttt{RG} denotes  Representation Guidance. Models at $400$K steps.}
    \centering
    \setlength{\tabcolsep}{4pt}
    \begin{tabular}{lccc}
    \toprule
    \Th{Model} & \Th{\#Params}  & \Th{FID$\downarrow$} \\ 
    \midrule 
    \texttt{DiT-B/2} w/ \texttt{ReDi}  & $130\text{M}$  & $25.7$ \\
    \texttt{DiT-B/2} w/ \texttt{ReDi}+ \texttt{RG}  & $130\text{M}$  & $20.2$ \\

    \midrule 
    \texttt{DiT-XL/2} w/ \texttt{ReDi}  & $675\text{M}$  & $8.7$ \\
    \texttt{DiT-XL/2} w/ \texttt{ReDi}+ \texttt{RG}  & $675\text{M}$  & $5.9$ \\
    \bottomrule
    \end{tabular}
    
    \label{tab:bench-rep-guid}

    \caption{\textbf{ReDi with REPA}. FID scores
on ImageNet 256×256 w/o CFG.}
    \centering
    \setlength{\tabcolsep}{4pt}
    \begin{tabular}{lccc}
    \toprule
    \Th{Model} & \Th{\#Iter.}  & \Th{FID$\downarrow$} \\ 
    \midrule 
    \texttt{SiT-XL/2} w/ \texttt{REPA}  & $4\text{M}$  & $5.9$ \\
   \rowcolor{TableColor} \texttt{SiT-XL/2} w/ \texttt{REPA}+\texttt{ReDi}  & $350\text{K}$  & $5.9$ \\

   \rowcolor{TableColor} \texttt{SiT-XL/2} w/ \texttt{REPA}+\texttt{ReDi}  & $1\text{M}$  & $3.5$ \\

    \bottomrule
    \end{tabular}

    \label{tab:bench-redi-repa}
\end{minipage}
\vspace{-10pt}

\end{table}

\begin{figure*}[t]
    \centering
    \setlength{\tabcolsep}{1pt} 
    \renewcommand{\arraystretch}{1.0} 
    \begin{tabular}{lcccccc}
        \rotatebox{90}{%
          \multirow{2}{*}{$\;\;\;\;\;\;$Image}%
        } & 
        \includegraphics[width=0.14\linewidth]{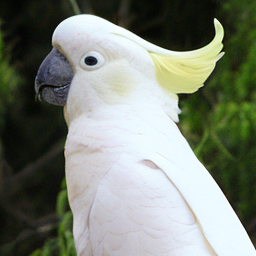} &
        \includegraphics[width=0.14\linewidth]{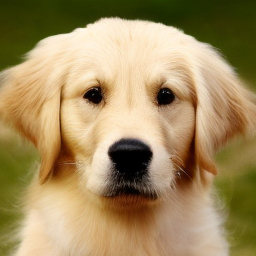} &
        \includegraphics[width=0.14\linewidth]{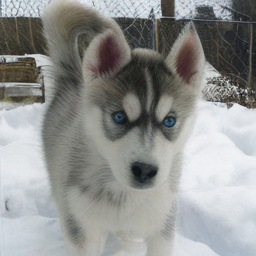} &
        \includegraphics[width=0.14\linewidth]{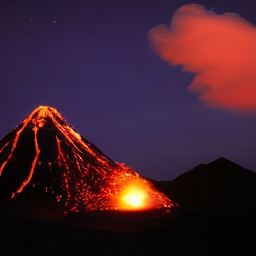} &
        \includegraphics[width=0.14\linewidth]{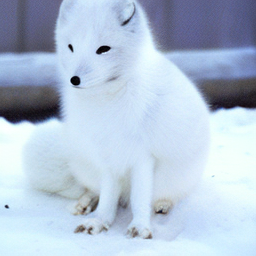} &
        \includegraphics[width=0.14\linewidth]{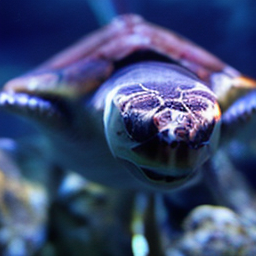} 
        \\

        \rotatebox{90}{%
          \begin{tabular}{@{}l@{}}
            \texttt{$\;\;\;\;\;\;$DINOv2}
          \end{tabular}%
        } & 
        \includegraphics[width=0.14\linewidth]{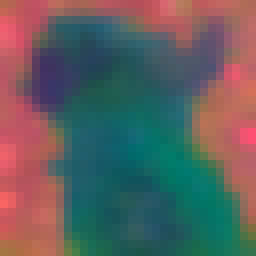} &
        \includegraphics[width=0.14\linewidth]{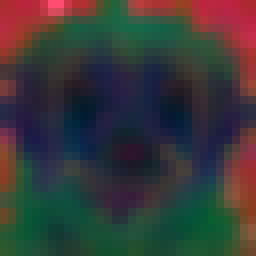} &
        \includegraphics[width=0.14\linewidth]{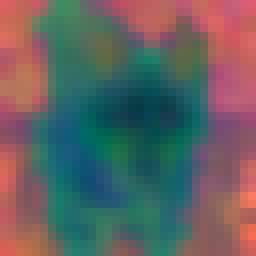} &
        \includegraphics[width=0.14\linewidth]{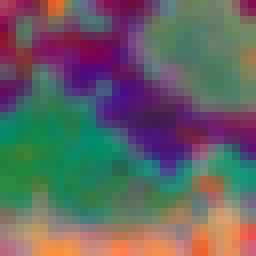} &
        \includegraphics[width=0.14\linewidth]{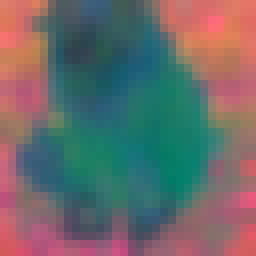} &
        \includegraphics[width=0.14\linewidth]{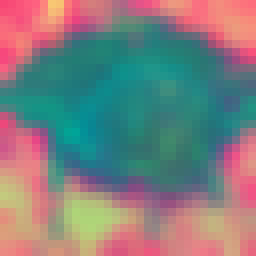} 
        \\
    \end{tabular}

    \caption{\textbf{Selected samples} from our \texttt{SiT-XL/2} w/ \texttt{ReDi} model trained on ImageNet $256\times256$. Images and visual representations are jointly generated by our model. We use Classifier-Free Guidance with $w = 4.0$.}
    \label{fig:qual}
    \vspace{-10pt}
\end{figure*}

\paragraph{Comparison with state-of-the-art generative models.} 
Ultimately, we provide a quantitative comparison between \texttt{ReDi} and other recent generative models using Classifier-Free Guidance (CFG) \citep{ho2022classifier} in \autoref{tab:sota_comparison}. Our method already outperforms both the vanilla \texttt{SiT-XL} and \texttt{SiT-XL} with \text{REPA} with only $350$ epochs. At $800$ epochs \texttt{ReDi} reaches an FID of $1.64$. We provide qualitative results of both generated images and visual representations in \autoref{fig:qual}.

\paragraph{Improving Unconditional Generation.} 
To establish the effectiveness of our method in improving generative models, we further present experiments for unconditional generation using \texttt{DiT}. As shown in \autoref{tab:bench-uncond}, our \texttt{ReDi} significantly improves generative performance for various model sizes. Specifically, with our \texttt{ReDi} FID drops from $69.3$ to $51.7$ for \texttt{B} and from $44.6$ to $25.1$ for \texttt{XL} models.

\subsection{Impact of Representation Guidance on generative performance.} 
\paragraph{Class Conditional Generation.}
In \autoref{tab:bench-rep-guid} we present the impact of Representation Guidance (\texttt{RG}) on generative performance. We observe that for both $\texttt{B}$ and $\texttt{XL}$ models, Representation Guidance unlocks further performance enhancements by guiding the generated image to closely follow the semantic features of $\texttt{DINOv2}$. Particularly for \texttt{DiT-XL} w/ \texttt{ReDi} the FID drops from $8.7$ to $5.9$. We also present qualitative results in \autoref{fig:rg_appendix}.

\paragraph{Unconditional Generation.} 
Representation Guidance is especially useful in unconditional generation scenarios, where the absence of class or text conditioning prevents the use of Classifier-Free Guidance to enhance performance. As demonstrated in \autoref{tab:bench-uncond}, Representation Guidance enhances the performance of \texttt{ReDi} with both \texttt{B} and \texttt{XL} models, \emph{further closing the performance gap between unconditional and conditional generation}.  Notably, \texttt{ReDi} with Representation Guidance achieves an FID of 22.6, approaching the performance of the class-conditioned \texttt{DiT-XL/2} (FID of 19.5).

\begin{figure}[tbp]
  \centering

  \begin{minipage}[]{0.48\textwidth}
    \centering
    \definecolor{vaeteal}{RGB}{0,128,128}
    \definecolor{classicorange}{RGB}{255,127,14}
    \vspace{30pt}
    \begin{tikzpicture}

      \begin{axis}[
        width=\textwidth,
        height=4.0cm,
        xlabel={\textbf{Classifier-Free Guidance weight} $\mathbf{w}$},
        ylabel={\textbf{FID}},
        grid=major,
        grid style={dashed,gray!30},
        legend pos=north east,
        mark size=3pt,
        tick label style={font=\small},
        label style={font=\small},
      ]
        \addplot+[
          vaeteal,
          thick,
          mark=o,
          smooth,
          tension=0.2,
          mark options={fill=white},
        ] coordinates {
          (1.35, 4.84)
          (1.50, 4.05)
          (2.00, 2.71)
          (2.40, 2.40)
          (2.50, 2.43)
          (2.65, 2.45)
          (2.85, 2.50)
          (3.00, 2.58)
          (3.50, 2.95)
          (4.00, 3.39)
          (4.50, 3.82)
        };
        \addlegendentry{\texttt{VAE}‑only CFG}

        \addplot+[
          classicorange,
          thick,
          mark=square*,
          smooth,
          tension=0.2,
          mark options={fill=white},
        ] coordinates {
          (1.10, 5.04)
          (1.20, 3.61)
          (1.30, 2.97)
          (1.35, 2.86)
          (1.40, 2.88)
          (1.50, 2.98)
          (1.60, 3.60)
          (1.70, 4.26)
        };
        \addlegendentry{\texttt{VAE} \string& \texttt{DINOv2} CFG}
      \end{axis}
    \end{tikzpicture}
    \caption{
    \textbf{VAE-only vs. VAE$\,\&\,$DINOv2 CFG.} FID scores for \texttt{SiT-XL} with \texttt{ReDi} (trained for 400K steps) as a function of Classifier-Free Guidance weight $w$, comparing two configurations: (1) applying CFG only to VAE latents (\texttt{VAE-only CFG}) versus (2) applying CFG to both VAE and \texttt{DINOv2} representations (\texttt{VAE$\,\&\,$DINOv2 CFG}).}
    \label{fig:cfg-ablation}
  \end{minipage}
  \hfill
  \begin{minipage}[]{0.49\textwidth}
    \centering
    \definecolor{barcolor}{RGB}{0,128,128}
    \begin{tikzpicture}
      \begin{axis}[
        ybar,
        width=\textwidth,
        height=4.0cm,
        bar width=9pt,
        ymin=24,
        ymax=47,
        ylabel={\textbf{FID}},
        xlabel={\textbf{\# Principal Components}},
        symbolic x coords={w/o \\ DINOv2,1,2,4,8,12,16,32},
        xtick={w/o \\ DINOv2,1,2,4,8,12,16,32},
        ymajorgrids=true,
        xticklabel style={},
        grid style=dashed,
        enlarge x limits=0.09,
        nodes near coords,
        every node near coord/.append style={
          font=\footnotesize,
          anchor=south,
          text=black
        },
        label style={font=\small},
        ticklabel style={font=\small},
      ]
        \addplot+[
          bar shift=0pt,
          fill=barcolor!50,
          draw=black
        ] coordinates {
          (1, 31.9)
          (2, 29.2)
          (4, 27.0)
          (8, 25.7)
          (12, 27.5)
          (16, 29.1)
          (32, 36.9)
        };

        \addplot+[
          bar shift=0pt,
          fill=gray!25,
          draw=black
        ] coordinates {
          (w/o \\ DINOv2, 43.0)
        };

        \draw[dashed, thick]
          (rel axis cs:0,0.826) -- (rel axis cs:1.0,0.826);
        \node[above right,font=\scriptsize]
          at (rel axis cs:0.735,0.82) {w/o~\texttt{DINOv2}};
      \end{axis}
    \end{tikzpicture}
    \vspace{-5pt}
    \caption{\textbf{Effect of number of principal components.} FID of \texttt{DiT-B/2} w/ \texttt{ReDi} with different number of \texttt{DINOv2} Principal Components. The vanilla \texttt{DiT-B/2} is illustrated with \textcolor{gray}{gray}. No Classifier-Free Guidance is used.}
    \label{fig:pca}
  \end{minipage}
  \vspace{-10pt}
\end{figure}
\begin{table}[t]

\caption{\textbf{Performance of Modality Combination Strategies.} \Th{FID} scores on ImageNet $256\times256$ without CFG for \texttt{DiT-B/2} with \texttt{ReDi} using Separate Tokens (\texttt{SP}) and Merged Tokens (\texttt{MR}). See \autoref{sec:additional_details} for details on throughput measurements.}

\centering
\setlength{\tabcolsep}{4pt}
\begin{tabular}{lccc}
\toprule
\Th{Model} & \Th{\#Tokens} & \Th{Throughput $\uparrow$}  & \Th{FID$\downarrow$} \\ 
\midrule 
 \rowcolor{TableColorGrey} \texttt{DiT-B/2}  & $256$ & $4.52$  & $43.5$ \\
 w/  \texttt{ReDi} (\texttt{MR})  & $256$ & $4.51$  & $25.7$ \\
  w/ \texttt{ReDi} (\texttt{SP})  &$512$ & $2.26$ & $24.7$ \\

\bottomrule
\end{tabular}
\vspace{3pt}
\label{tab:aggr-ablation}
\vspace{-15pt}
\end{table}

\subsection{Analysis}

\paragraph{Dimensionality reduction ablation.} We begin the analysis of our method by ablating the impact of dimensionality reduction on the visual representations, as shown in \autoref{fig:pca}. Initially, we observe that jointly learning as little as one principal component yields significant improvements in generative performance.  Increasing the component count continues to improve performance, up to $r=8$, beyond which further components begin to degrade the quality of generation. This suggests an optimal intermediate subspace where compressed visual features retain sufficient expressivity to guide generation without dominating model capacity.

\paragraph{Merged Tokens vs. Separate Tokens.} 
In \autoref{tab:aggr-ablation}, we evaluate the effectiveness of the two explored integration strategies, Merged Tokens (\texttt{MR}) and Separate Tokens (\texttt{SP}), for joint learning of image VAE latents and visual representations, using \texttt{DiT-B/2} as our base model. While both approaches achieve comparable performance gains, \texttt{SP} demonstrates slightly better results. This advantage comes at a significant computational cost: \texttt{SP} doubles the transformer's input sequence length by introducing $256$ additional \texttt{DINOv2} tokens, resulting in approximately $2\times$ greater compute demands during both training and inference \citep{kaplan2020scaling}. The \texttt{MR} strategy, by contrast, maintains the original sequence length while delivering similar performance improvements, thereby preserving computational efficiency as measured by throughput.

\paragraph{VAE-only Classifier-Free Guidance.}
As \texttt{ReDi} jointly models both VAE latents and visual representations, we investigate two Classifier-Free Guidance (CFG) strategies: applying CFG exclusively to VAE latents (\texttt{VAE-only CFG}) versus applying it to both modalities simultaneously (\texttt{VAE$\,\&\,$DINOv2 CFG}). Our experiments in \autoref{fig:cfg-ablation} demonstrate that VAE-only CFG achieves superior results, yielding an FID of $2.39$ compared to $2.86$ for the \texttt{VAE$\,\&\,$DINOv2 CFG} approach. Notably, \texttt{VAE-only CFG} also shows greater robustness to variations in the CFG weight parameter.

\section{Conclusion}
\label{sec:conclusion}

In this work, we explore the relationship between semantic representation learning and generative performance in latent diffusion models. Building on recent insights, we introduced \texttt{ReDi}, a novel framework that integrates high-level semantic features with low-level latent representations within the diffusion process. Unlike prior approaches that rely on auxiliary objectives, \texttt{ReDi} jointly models the two distributions. We demonstrate that this simple approach is more effective at leveraging the semantic features and leads to drastic improvements in generative performance.
We further proposed Representation Guidance, a novel guidance method that leverages the jointly learned semantic features to enhance image quality. Across both conditional and unconditional settings, \texttt{ReDi} consistently improves generation quality and accelerates convergence, highlighting the benefits of our approach. 

\paragraph{Acknowledgements}

This work has been partially supported by project MIS 5154714 of the National Recovery and Resilience Plan Greece 2.0 funded by the European Union under the NextGenerationEU Program and by Institute of Informatics and Telecommunications, National Center for Scientific Research “Demokritos”. 
Hardware resources were granted with the support of GRNET. Also, this work was performed using HPC resources from GENCI-IDRIS (Grants 2024-AD011012884R3).

\addtocontents{toc}{\protect\setcounter{tocdepth}{2}}

\bibliography{main}
\bibliographystyle{icml2025}


\newpage

\appendix

\newpage

{\huge \textbf{Appendix}}

\renewcommand{\thesection}{\Alph{section}}
{
  \hypersetup{linkcolor=black}
  \tableofcontents
}




\section{ReDi with Stochastic Interpolant Models (SiT)}
\label{sec:sit_redi}
In the main paper, we introduced \texttt{ReDi} within the DDPM framework, as employed by \texttt{DiT} models. In this section, we begin with a brief overview of Stochastic Interpolant Models \cite{ma2024sit} and then describe how \texttt{ReDi} can be applied in this setting.

\subsection{Stochastic Interpolant Models (SiT)}
Following flow-based models \cite{lipman2023flow}, stochastic interpolants involve a continuous time-dependent process transforming a data distribution $\mathbf{x_0} \sim p(\mathbf{x})$ into Gaussian noise $\boldsymbol{\epsilon} \sim \mathcal{N}(\mathbf{0}, \textbf{I})$:

\begin{equation}
    \label{eq+:sit-forward}
    \mathbf{x}_t = \alpha_t \mathbf{x}_0 + \sigma_t \boldsymbol{\epsilon}, \quad \alpha_0 = \sigma_1 = 1, \quad \alpha_1 = \sigma_0 = 0,
\end{equation}
where $\alpha_t$ and $\sigma_t$ are increasing and decreasing functions of $t$ respectively.

 Given this process, the marginal probability distribution $p_t(\mathbf{x})$ of $\mathbf{x}_t$ in (\ref{eq+:sit-forward})
coincides with the distribution of the probability flow ordinary differential equation  with a velocity field: 
\begin{equation}
    \label{eq:vel_field}
    \mathbf{\dot{x}}_t = \mathbf{v}(\mathbf{x}_t, t).
\end{equation}
The velocity field can be approximated by a neural network $\mathbf{v}_\theta(x_t, t)$ by minimizing the following training objective:
\begin{equation}
    \mathcal{L}_{\mathrm{velocity}}(\theta)
\;:=\;
\mathbb{E}_{\mathbf{x}_0, \epsilon, t}
\Bigl\|\,
\mathbf{v}_{\theta}(\mathbf{x}_t, t)
\;-\;
\dot{\alpha}_t\,\mathbf{x}_0
\;-\;
\dot{\sigma}_t\,\boldsymbol{\epsilon}
\,\Bigr\|^2
\,.
\end{equation}

\subsection{Joint Image-Representation Generation with SiT}
During training, given a VAE latent image $\mathbf{x}_0$ and a visual representation $\mathbf{z_0}$, we define a joint interpolation process:
\begin{equation}
\label{eq:joint_forward_proc_sit}  
    \mathbf{x}_t = \alpha_t \mathbf{x}_0 + \sigma_t \boldsymbol{\epsilon}_x, \quad  
    \mathbf{z}_t = \alpha_t \mathbf{z}_0 + \sigma_t \boldsymbol{\epsilon}_z, 
\end{equation}  

The model $\mathbf{v_\theta}(\mathbf{x}_t, \mathbf{z}_t, t)$ takes as input $\mathbf{x}_t$ and $\mathbf{z}_t$, along with timestep $t$, and jointly predicts the velocity for both inputs. Specifically, it produces two separate predictions: $\textcolor{black}{\mathbf{v}^x_\theta}(\mathbf{x}_t, \mathbf{z}_t, t)$ for the image latent velocity $\mathbf{v}_x$,  and $\textcolor{black}{\mathbf{v}^z_\theta}(\mathbf{x}_t, \mathbf{z}_t, t)$ for the visual representation velocity $\mathbf{v}_z$. The training objective combines both predictions:
\begin{equation}  
    \label{eq:joint_objective_sit}  
    \mathcal{L}_{joint} = \underset{\mathbf{x_0}, \mathbf{z_0}, t} { \mathbb{E}} \Big [  
    \Vert \textcolor{black}{\mathbf{v}^x_\theta}(\mathbf{x}_t,\mathbf{z}_t, t) - 
    \dot{\alpha}_t\,\mathbf{x}_0 -
    \dot{\sigma}_t\,\boldsymbol{\epsilon}_x
    \Vert^2  
    + \lambda_z \Vert \textcolor{black}{\mathbf{v}^z_\theta}(\mathbf{x}_t,\mathbf{z}_t, t) - 
    \dot{\alpha}_t\,\mathbf{z}_0 -
    \dot{\sigma}_t\,\boldsymbol{\epsilon}_z
    \Vert^2 \Big],  
\end{equation}  
where $\lambda_z$ balances the velocity loss for $\mathbf{z}_t$. By default, we use $\lambda_z = 1$, $\alpha_t = t$ and $ \sigma_t = 1-t$ in our experiments.

\section{Additional Implementation Details}
\label{sec:additional_details}
\subsection{Architecture details}
We present in \autoref{tab:model-sizes} the configurations of the different-sized \texttt{DiT} and \texttt{SiT} models used in our experiments.

\begin{table}[h]

\caption{\textbf{Model configuration details.} The configurations are the same for both \texttt{DiT} and \texttt{SiT} models.}
\centering
\setlength{\tabcolsep}{4pt}
\begin{tabular}{lccc}
\toprule
\Th{Model Size} & \Th{B/2}  & \Th{L/2} & \Th{XL/2} \\ 
\midrule 
  \texttt{Input Size} & $32\times32\times4$  & $32\times32\times4$ & $32\times32\times4$ \\

  \texttt{Patch Size} & $2$  & $2$ & $2$ \\

  \texttt{\# Layers} & $12$  & $24$ & $28$ \\

  \texttt{\# Heads} & $12$  & $16$ & $16$ \\

  \texttt{Hidden Dim.} & $768$  & $1024$ & $1152$ \\

\bottomrule
\end{tabular}
\vspace{3pt}

\label{tab:model-sizes}
\end{table}

\subsection{Optimization details}
We present in \autoref{tab:opt-details} the optimization hyperparameters used for all experiments presented in the paper.

\begin{table}[h]

\caption{\textbf{Optimization details.} The optimization hyperparameters for both \texttt{DiT} and \texttt{SiT} models.}
\centering
\setlength{\tabcolsep}{4pt}
\begin{tabular}{cc}
\toprule
\texttt{Batch Size} & $256$   \\ 
\texttt{Optimizer} & \texttt{AdamW}   \\ 
\texttt{LR} & $10^{-4}$   \\ 
$(\beta_1, \beta_2)$ & $(0.9, 0.999)$   \\

\bottomrule
\end{tabular}

\label{tab:opt-details}
\end{table}

\paragraph{Computational Resources.} For both training and sampling we use 8 NVIDIA A$100$ $40$GB GPUs. Throughput, as presented in \autoref{tab:aggr-ablation} is measured on a single NVIDIA A$100$ $40$GB GPU with a batch size of 64 as the number of images generated per second using $250$ sampling steps.

\subsection{Further implementation details}
\label{sec:further_impl}

\paragraph{ReDi with REPA experiment.} To apply the Representation Alignment objective (\texttt{REPA}) on top of \texttt{ReDi} we follow the implementation of \citep{Yu2025repa} and employ a 
 projection layer in the $8{\text{th}}$ transformer layer. The projection is a three-layer MLP with SiLU activations \citep{elfwing2018sigmoid}. The weight on alignment loss is $\lambda_{\texttt{REPA}}=0.5$.

\section{Detailed Benchmarks}

We provide a detailed evaluation of the main experiments presented in the main paper, including additional metrics and training iterations.  
 Specifically, ~\autoref{tab:details_redi} details the performance of the \texttt{SiT-XL/2} w/ \texttt{ReDi}  models. Further ~\autoref{tab:details_redirepa}  presents results for the \texttt{ReDi} with \texttt{REPA} (\texttt{SiT-XL/2}). For all models, we use the evaluation metrics reported in the original publications.

\begin{table}[!h]
\centering
\setlength{\tabcolsep}{4pt}
\begin{tabular}{lrrrrrrr}
\toprule
\Th{Model} & \Th{\#Iters.} &  \Th{FID$\downarrow$} & \Th{sFID$\downarrow$} & \Th{IS$\uparrow$} & \Th{Prec.$\uparrow$} & \Th{Rec.$\uparrow$} \\
\midrule

\rowcolor{TableColorGrey}  \texttt{SiT-XL/2} 
\citet{peebles2023scalable} & $7\text{M}$ & $8.3$ & $6.3$  & $131.7$ & $0.68$ & $0.67$ \\

w/ \our &  $50\text{K}$  & $56.1$ & $18.9$ \hspace{0.05cm} & $23.8$ & $0.44$ & $0.47$ \\
w/ \our &  $100\text{K}$ & $23.1$ & $5.9$  & $61.5$ & $0.64$ & $0.57$ \\
w/ \our &  $200\text{K}$ & $12.6$ & $5.7$  & $97.3$ & $0.69$ & $0.61$ \\
w/ \our &  $300\text{K}$ & $9.7$ & $5.3$  & $117.3$ & $0.71$ & $0.62$ \\
w/ \our &  $400\text{K}$ & $7.5$ & $5.1$  & $129.5$ & $0.72$ & $0.62$ \\
w/ \our &  $4\text{M}$ & $3.3$ & $4.8$  & $188.9$ & $0.74$ & $0.68$ \\

\bottomrule
\end{tabular}
\vspace{3pt}
\caption{\textbf{Detailed evaluation} for \texttt{SiT-XL/2} w/ \texttt{ReDi}. All results are reported without classifier-free guidance.}
\label{tab:details_redi}
\end{table}

\begin{table}[!h]
\centering
\setlength{\tabcolsep}{4pt}
\begin{tabular}{lrrrrrrr}
\toprule
\Th{Model} & \Th{\#Iters.} &  \Th{FID$\downarrow$} & \Th{sFID$\downarrow$} & \Th{IS$\uparrow$} & \Th{Prec.$\uparrow$} & \Th{Rec.$\uparrow$} \\
\midrule

\rowcolor{TableColorGrey}  \texttt{SiT-REPA-XL/2} 
\citet{Yu2025repa} & $400\text{K}$ & $7.9$ & $5.1$  & $122.6$ & $0.70$ & $0.65$ \\

\rowcolor{TableColorGrey}  \texttt{SiT-REPA-XL/2} 
 & $4\text{M}$ & $5.9$ & $ 5.7$  & $ 157.8$ & $0.70$ & $0.69$ \\

w/ \our &  $50\text{K}$  & $44.8$ & $18.7$ \hspace{0.05cm} & $32.8$ & $0.50$ & $0.49$ \\
w/ \our &  $100\text{K}$ & $15.2$ & $5.6$  & $85.3$ & $0.68$ & $0.59$ \\
w/ \our &  $200\text{K}$ & $8.3$ & $5.2$  & $122.3$ & $0.71$ & $0.61$ \\
w/ \our &  $300\text{K}$ & $6.3$ & $5.1$  & $140.6$ & $0.73$ & $0.62$ \\
w/ \our &  $400\text{K}$ & $5.3$ & $4.9$  & $149.8$ & $0.74$ & $0.63$ \\
w/ \our &  $1\text{M}$ & $3.5$ & $4.64$  & $177.9$ & $0.75$ & $0.69$ \\

\bottomrule
\end{tabular}
\vspace{3pt}
\caption{\textbf{Detailed evaluation} for \texttt{ReDi} with \texttt{REPA}.  All results are reported without classifier-free guidance.}
\label{tab:details_redirepa}
\end{table}

\section{Baseline Generative Models}
We provide here a brief description of the baseline approaches presented in the main paper. Specifically, we consider 
(a) \emph{Autoregressive Models}, (b) \emph{Latent Diffusion Models}, and (c) \texttt{REPA} \citep{Yu2025repa} that also \emph{leverages visual representations} to enhance generative performance.

\paragraph{(a) Autoregressive Models}
\begin{itemize}
    \item \texttt{VAR} \citep{tian2024visual}
proposes a scalable generative framework that autoregressively predicts higher-resolution image details from lower-resolution contexts across multiple scales.

    \item \texttt{MagViTv2} \citep{yu2024language} introduces a lookup-free quantization method enabling  a large vocabulary
   that is able to improve the generation quality of autoregressive models.

    \item \texttt{MAR} \citep{li2024autoregressive} proposes an autoregressive image generation framework that eliminates the need for vector quantization

\end{itemize}

\paragraph{(b) Latent Diffusion Models}

\begin{itemize}
\item \texttt{LDM} \citep{rombach2022high} proposes latent diffusion models, modeling the image distribution in a compressed latent
space produced by a KL- or VQ-regularized autoencoder.

\item \texttt{U-ViT-H/2} \cite{bao2023all} proposes a \texttt{ViT}-based \citep{dosovitskiy2021image} latent diffusion model that incorporates skip connections.

\item \texttt{DiT} \cite{peebles2023scalable} proposes a pure transformer backbone for training diffusion models and incorporates AdaIN-zero
modules.

\item \texttt{MaskDiT} \citep{zheng2023fast} trains diffusion transformers with an auxiliary mask reconstruction task

\item \texttt{MDT} \cite{gao2023mdtv2} introduce an effective
mask latent modeling scheme, and design an asymmetric masking diffusion transformer.

\item \texttt{SD-DiT} \citep{zhu2024sd} extends the MaskDiT architecture by incorporating a discrimination objective using a
momentum encoder.

\item \texttt{SiT} \citep{ma2024sit}  improves diffusion transformer training by moving from discrete diffusion to continuous flow-based modeling.

\item \texttt{FasterDiT} \citep{yao2024fasterdit}  incorporates supervision of the velocity direction into the denoising objective, significantly accelerating the training process.

\end{itemize}

\paragraph{(c) Leveraging Visual Representations}

\begin{itemize}
    \item \texttt{REPA} \citep{Yu2025repa} aligns the representations of diffusion transformer models to the representations of self-supervised models.
\end{itemize}

\section{Limitations \& Future Work}
\label{sec:limitations_and_feature}
This section outlines some limitations of our current work and highlights promising directions for future research.

\paragraph{Multiple visual representations.} In this work, we demonstrate the effectiveness of jointly modeling the visual representations from \texttt{DINOv2} during the diffusion process. A promising direction for future research is to investigate whether integrating \emph{multiple} visual representations, each capturing different semantic or structural properties, can further boost generative performance.

\paragraph{Different dimensionality reduction approaches.}We have shown that projecting visual representations into a lower-dimensional space with PCA effectively compresses visual features while retaining sufficient information. An interesting direction for future work is to explore more sophisticated compression techniques, such as training an autoencoder, to better capture and retain the expressivity of these features.

\section{Broader Impact}
\label{sec:soc_impact}
Generative models carry a substantial risk of misuse. Their application can lead to various negative societal impacts, most notably the spread of disinformation. Enhancements in generative performance, as achieved by our method, may further increase the realism of generated content, potentially making disinformation even more convincing.

\newpage
\section{Additional Qualitative Results}

\begin{figure*}[!htbp]
    \centering
    \setlength{\tabcolsep}{1pt} 
    \renewcommand{\arraystretch}{1.0} 
    \begin{tabular}{ccccccc}
        wo/ \texttt{RG} & $w_r = 1.1$ & $w_r=1.2$ & 
        $w_r=1.3$ & $w_r=1.4$ & $w_r=1.5$ & $w_r=1.6$ \\
        
        \includegraphics[width=0.13\linewidth]{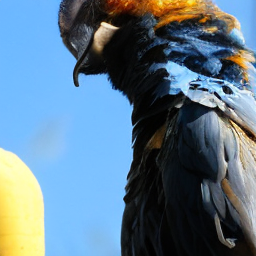} &
        \includegraphics[width=0.13\linewidth]{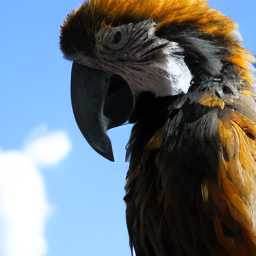} &
        \includegraphics[width=0.13\linewidth]{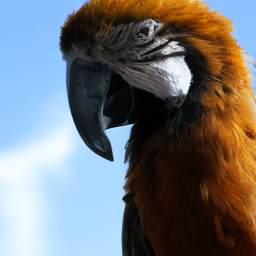} &
        \includegraphics[width=0.13\linewidth]{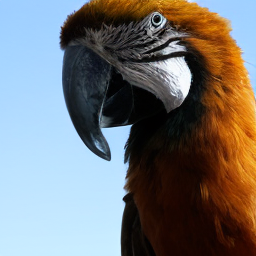} &
        \includegraphics[width=0.13\linewidth]{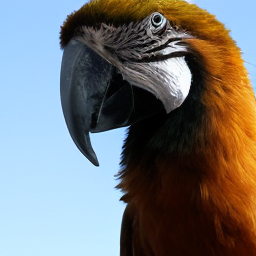} &
        \includegraphics[width=0.13\linewidth]{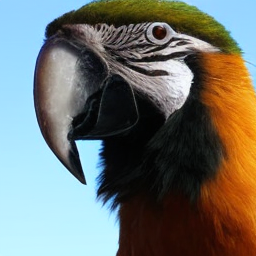} &
        \includegraphics[width=0.13\linewidth]{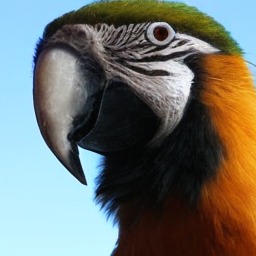} 
        \\


        \includegraphics[width=0.13\linewidth]{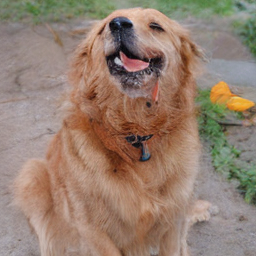} &
        \includegraphics[width=0.13\linewidth]{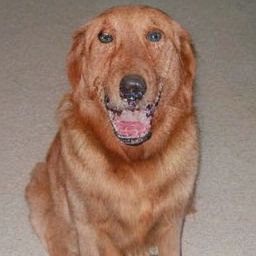} &
        \includegraphics[width=0.13\linewidth]{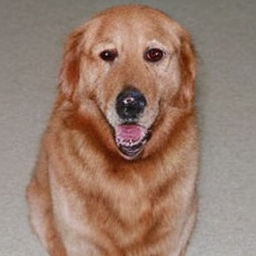} &
        \includegraphics[width=0.13\linewidth]{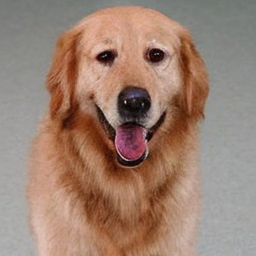} &
        \includegraphics[width=0.13\linewidth]{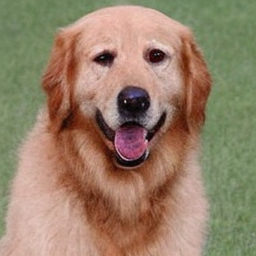} &
        \includegraphics[width=0.13\linewidth]{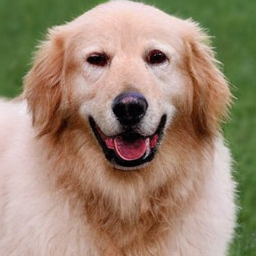} &
        \includegraphics[width=0.13\linewidth]{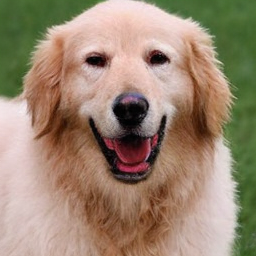} 
        \\

        \includegraphics[width=0.13\linewidth]{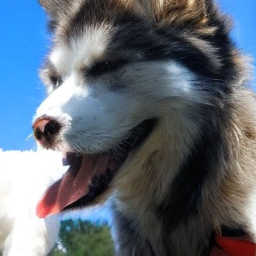} &
        \includegraphics[width=0.13\linewidth]{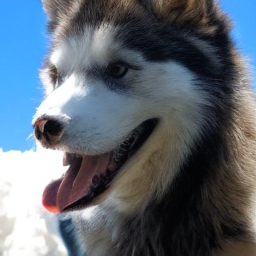} &
        \includegraphics[width=0.13\linewidth]{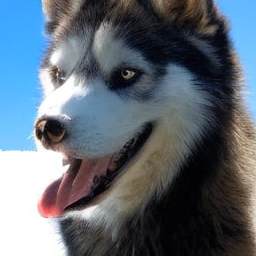} &
        \includegraphics[width=0.13\linewidth]{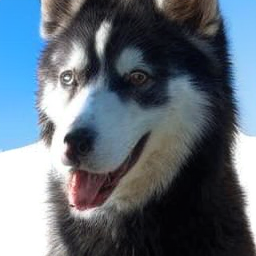} &
        \includegraphics[width=0.13\linewidth]{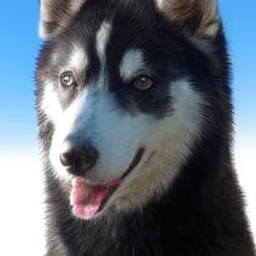} &
        \includegraphics[width=0.13\linewidth]{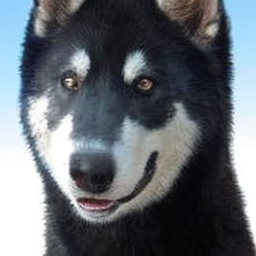} &
        \includegraphics[width=0.13\linewidth]{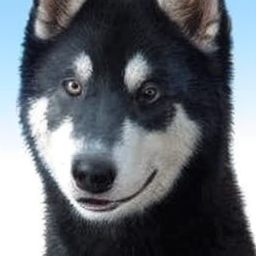} 
        \\
        
        \includegraphics[width=0.13\linewidth]{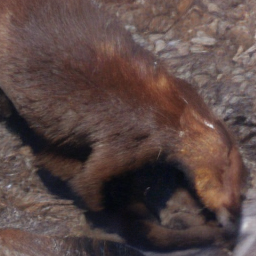} &
        \includegraphics[width=0.13\linewidth]{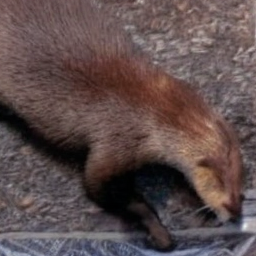} &
        \includegraphics[width=0.13\linewidth]{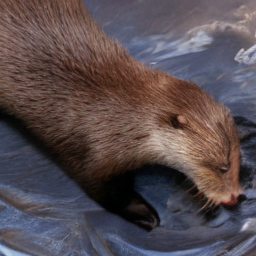} &
        \includegraphics[width=0.13\linewidth]{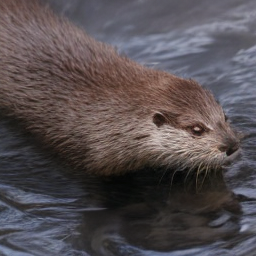} &
        \includegraphics[width=0.13\linewidth]{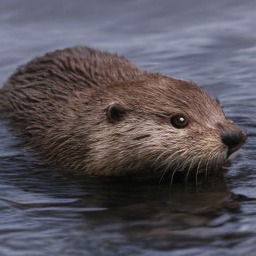} &
        \includegraphics[width=0.13\linewidth]{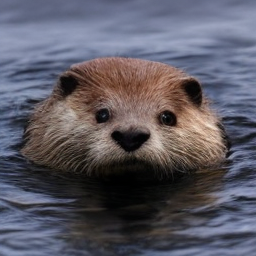} &
        \includegraphics[width=0.13\linewidth]{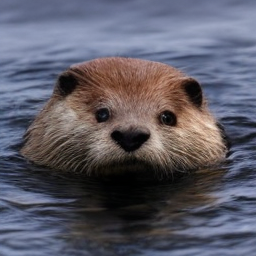} 
        \\

        \includegraphics[width=0.13\linewidth]{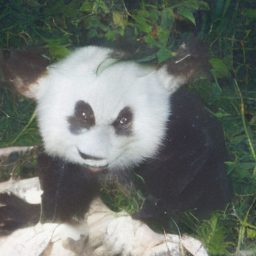} &
        \includegraphics[width=0.13\linewidth]{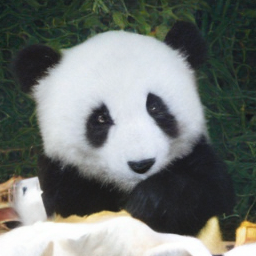} &
        \includegraphics[width=0.13\linewidth]{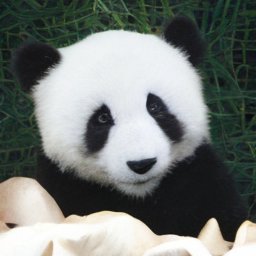} &
        \includegraphics[width=0.13\linewidth]{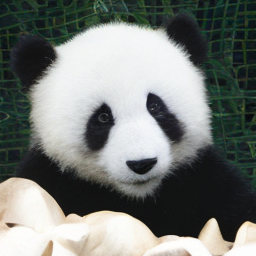} &
        \includegraphics[width=0.13\linewidth]{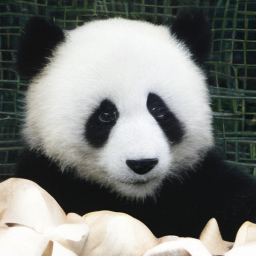} &
        \includegraphics[width=0.13\linewidth]{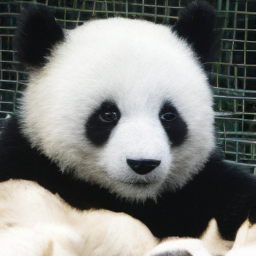} &
        \includegraphics[width=0.13\linewidth]{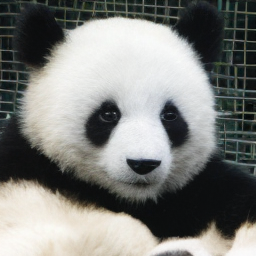} 
        \\

        \includegraphics[width=0.13\linewidth]{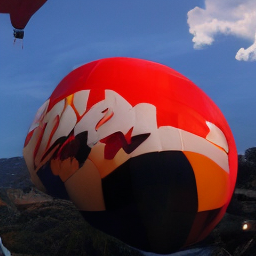} &
        \includegraphics[width=0.13\linewidth]{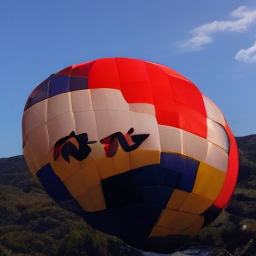} &
        \includegraphics[width=0.13\linewidth]{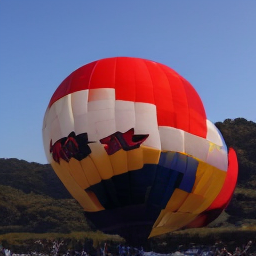} &
        \includegraphics[width=0.13\linewidth]{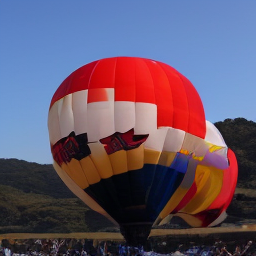} &
        \includegraphics[width=0.13\linewidth]{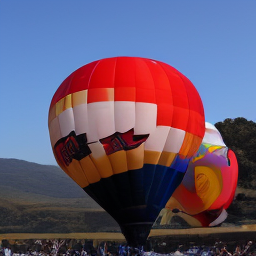} &
        \includegraphics[width=0.13\linewidth]{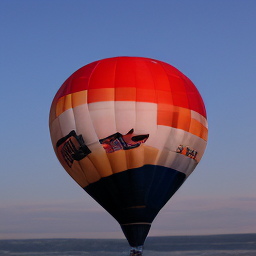} &
        \includegraphics[width=0.13\linewidth]{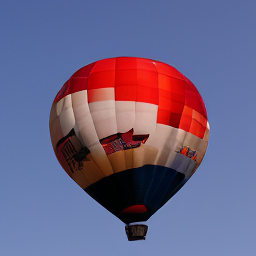} 
        \\


    \end{tabular}

    \caption{\textbf{The effect of Representation Guidance.} Samples from our \texttt{DiT-XL/2} w/ \texttt{ReDi} model trained on ImageNet $256\times256$ for $400$k steps with different Representation Guidance weights $w_r$.}
    \label{fig:rg_appendix}
    \vspace{-10pt}
\end{figure*}

\begin{figure}[!htbp]
        \centering

        \includegraphics[width=0.9\textwidth]{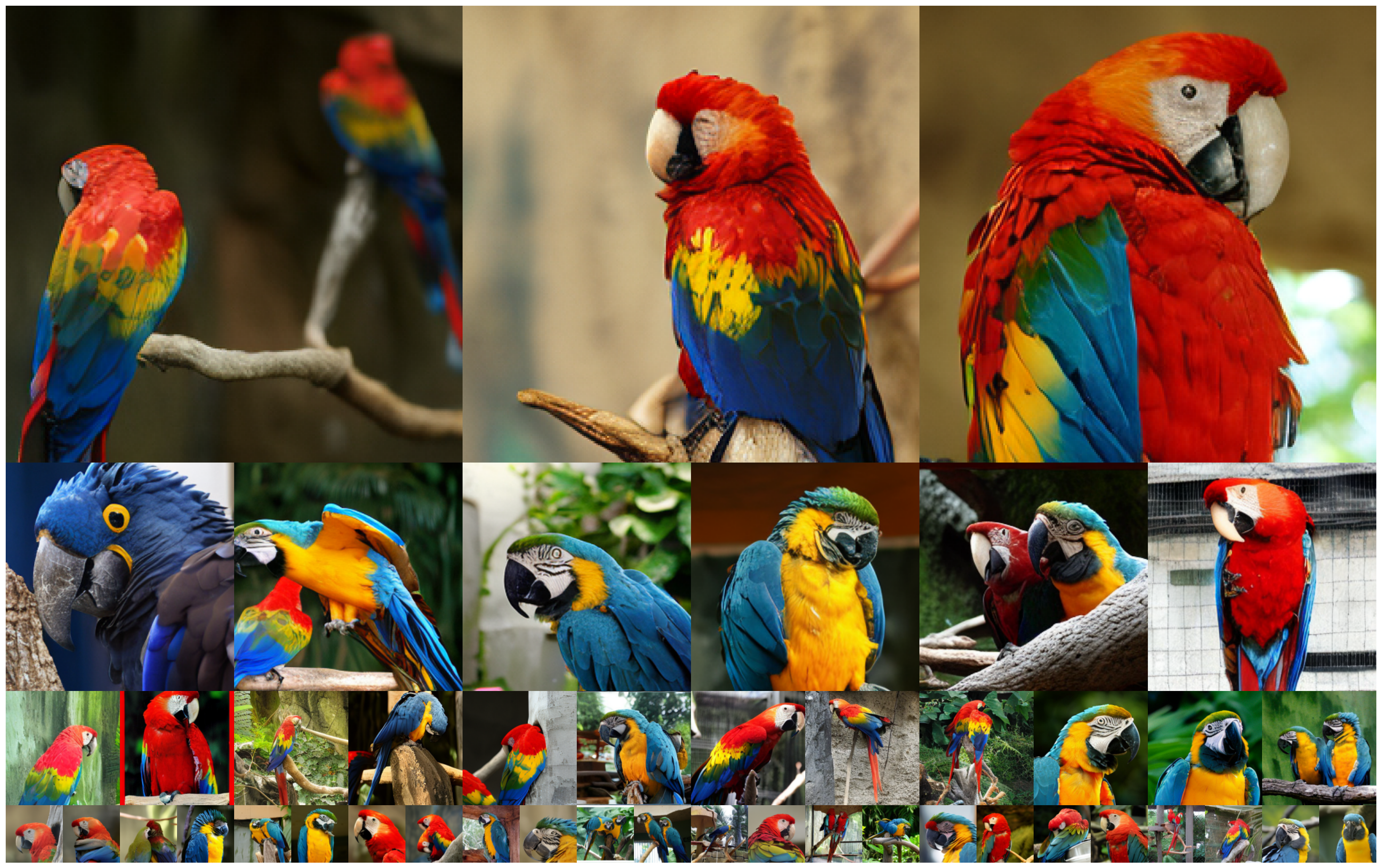} 
    \caption{\textbf{Uncurated generation results} of \texttt{SiT-XL/2} w/ \texttt{ReDi}. We use Classifier-Free Guidance
with $w = 4.0$. Class label = $88$.}
    \label{fig:uncurated_samples_1}
\end{figure}

\begin{figure}[!htbp]
        \centering

        \includegraphics[width=0.9\textwidth]{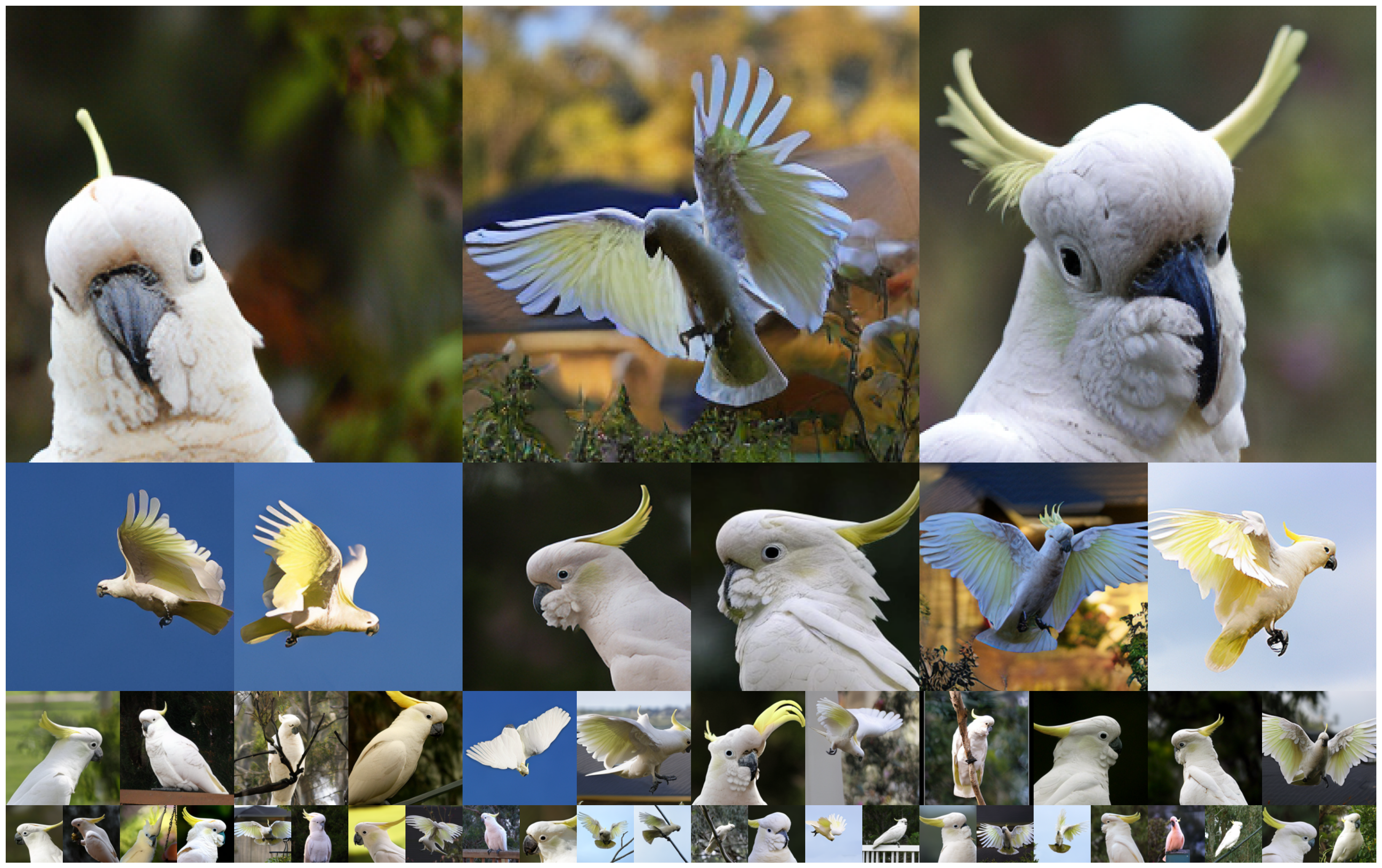} 
    \caption{\textbf{Uncurated generation results} of \texttt{SiT-XL/2} w/ \texttt{ReDi}. We use Classifier-Free Guidance
with $w = 4.0$. Class label = $89$.}
    \label{fig:uncurated_samples_1}
\end{figure}

\begin{figure}[!htbp]
        \centering

        \includegraphics[width=0.9\textwidth]{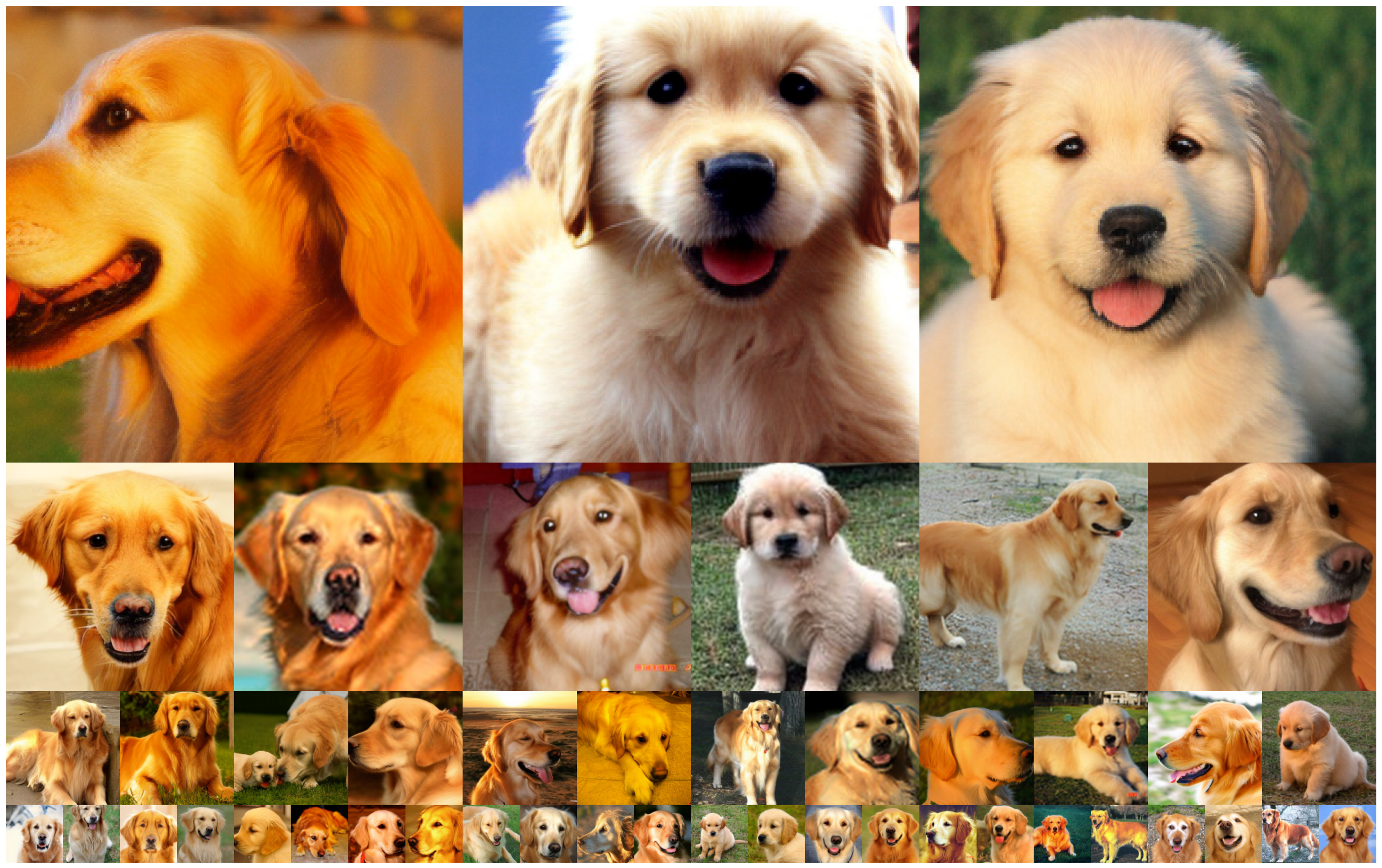} 
    \caption{\textbf{Uncurated generation results} of \texttt{SiT-XL/2} w/ \texttt{ReDi}. We use Classifier-Free Guidance
with $w = 4.0$. Class label = $207$.}
    \label{fig:uncurated_samples_1}
\end{figure}

\begin{figure}[!htbp]
        \centering

        \includegraphics[width=0.9\textwidth]{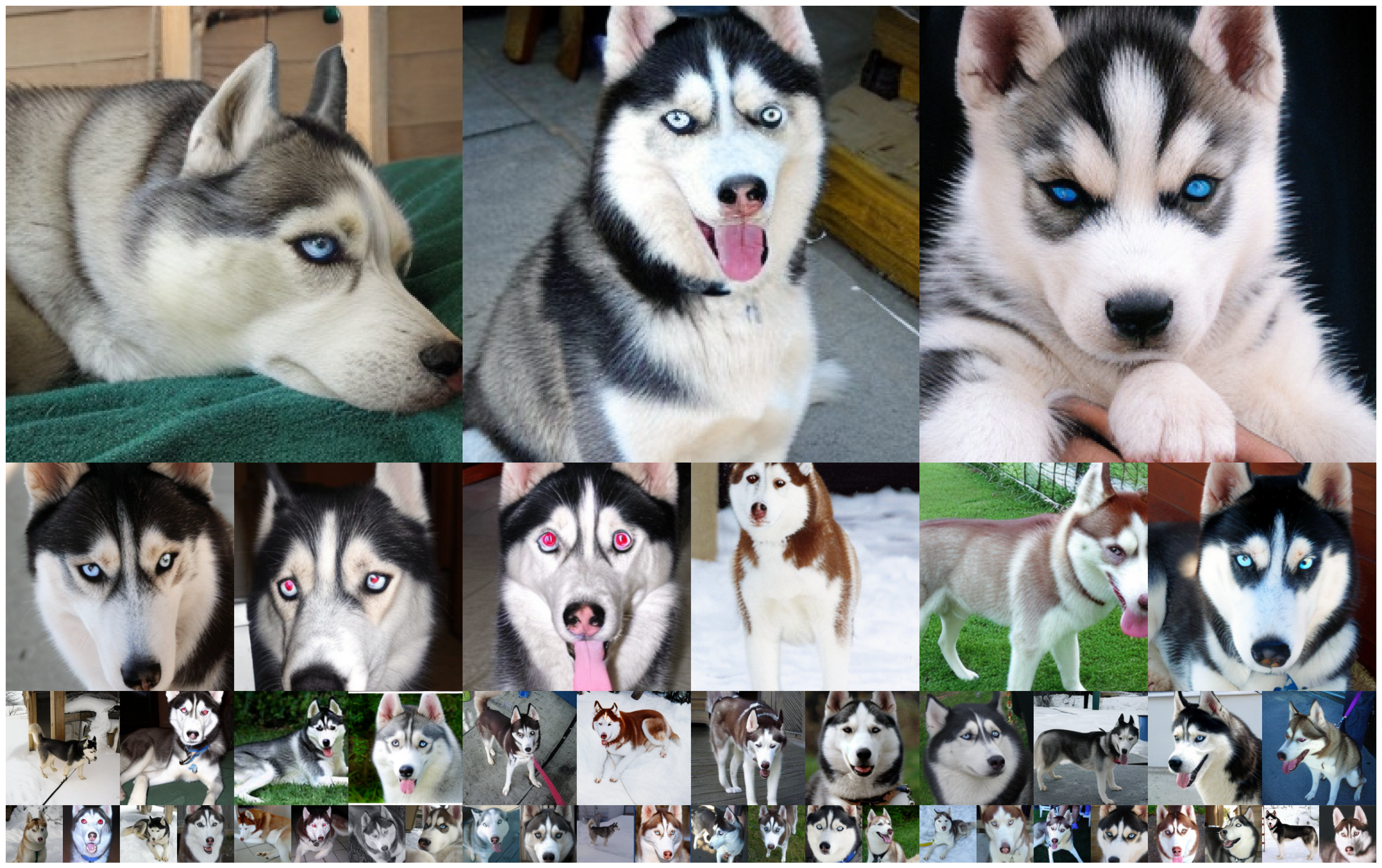} 
    \caption{\textbf{Uncurated generation results} of \texttt{SiT-XL/2} w/ \texttt{ReDi}. We use Classifier-Free Guidance
with $w = 4.0$. Class label = $250$.}
    \label{fig:uncurated_samples_1}
\end{figure}

\begin{figure}[!htbp]
        \centering

        \includegraphics[width=0.9\textwidth]{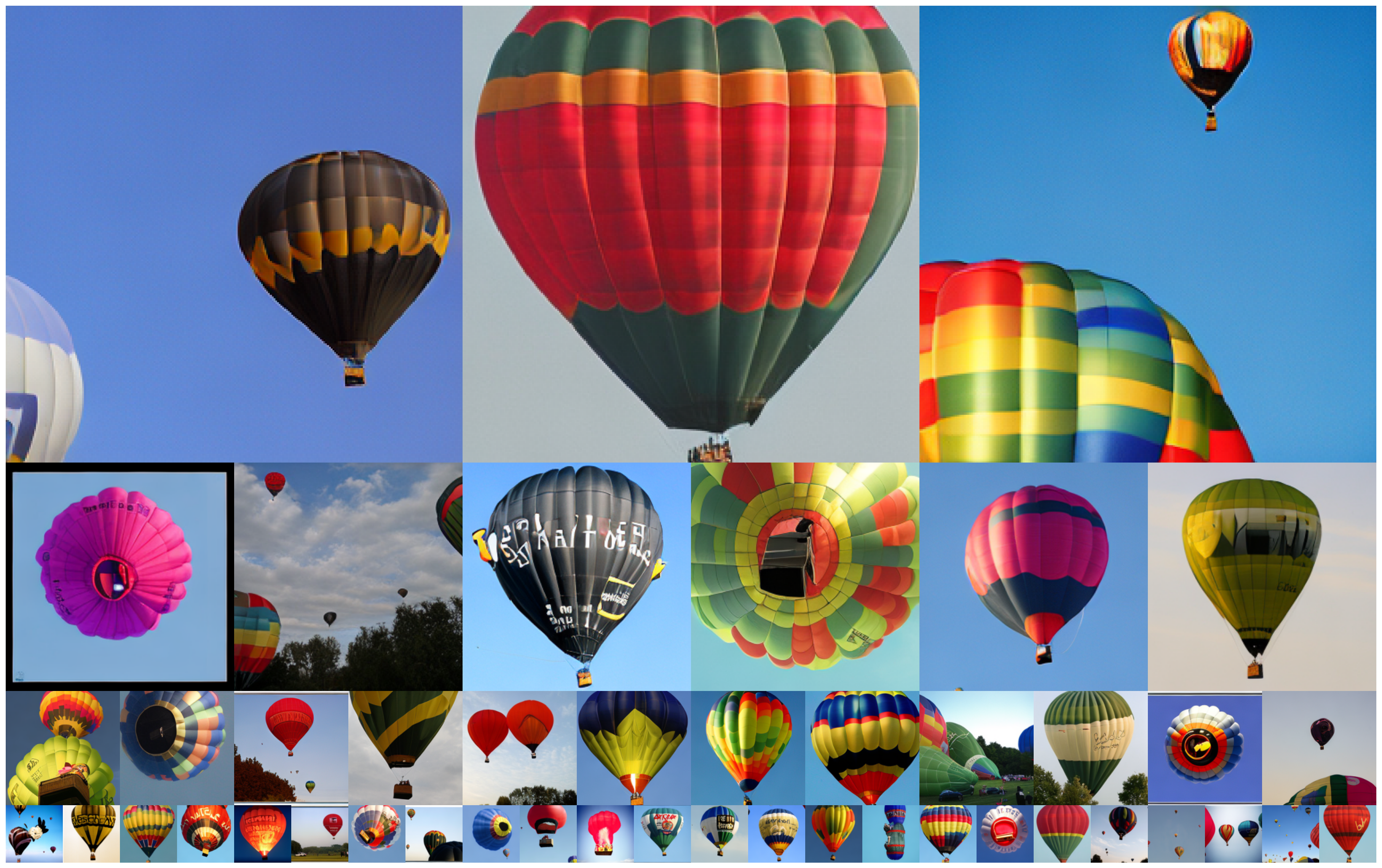} 
    \caption{\textbf{Uncurated generation results} of \texttt{SiT-XL/2} w/ \texttt{ReDi}. We use Classifier-Free Guidance
with $w = 4.0$. Class label = $417$.}
    \label{fig:uncurated_samples_1}
\end{figure}

\begin{figure}[!htbp]
        \centering

        \includegraphics[width=0.9\textwidth]{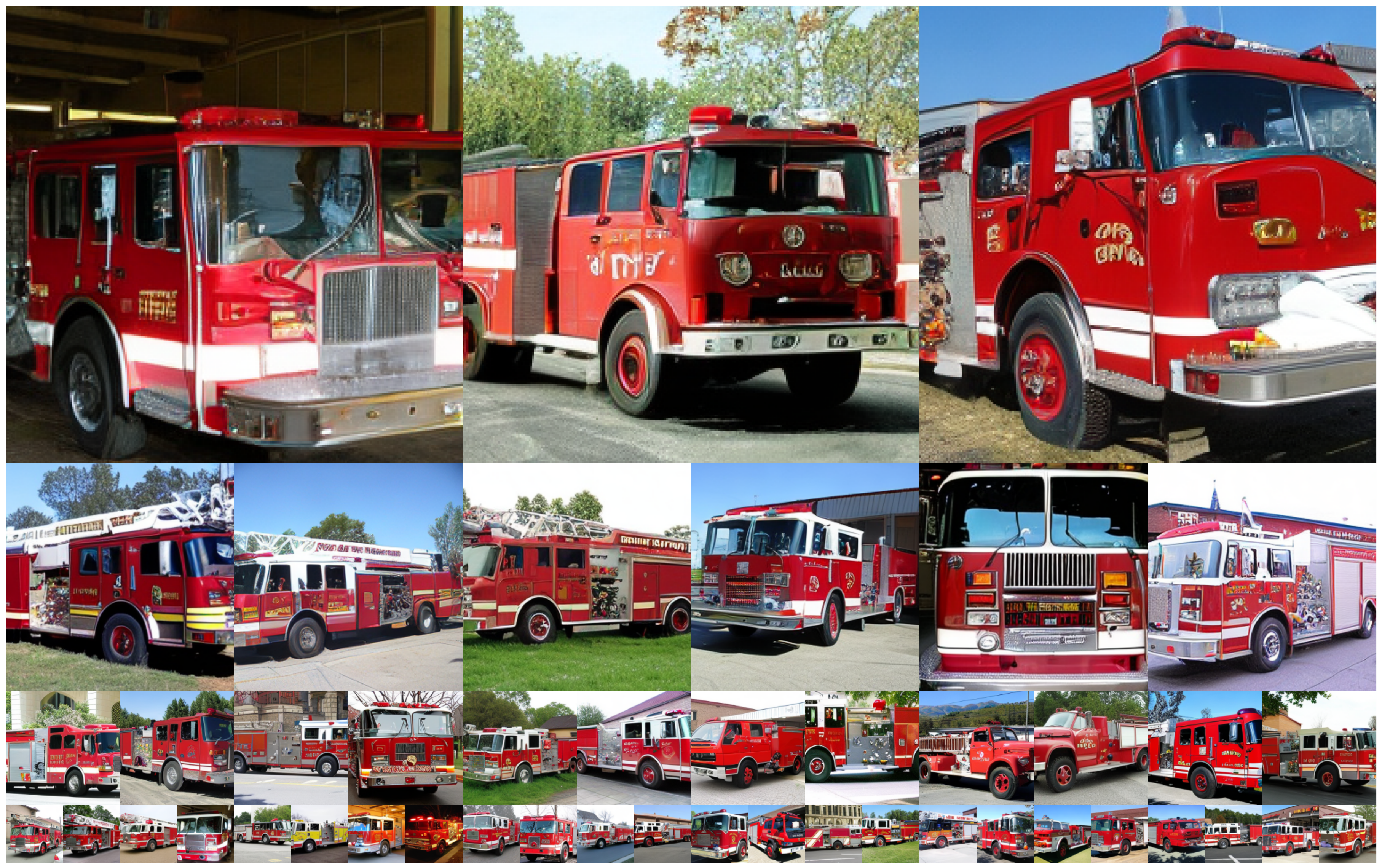} 
    \caption{\textbf{Uncurated generation results} of \texttt{SiT-XL/2} w/ \texttt{ReDi}. We use Classifier-Free Guidance
with $w = 4.0$. Class label = $555$.}
    \label{fig:uncurated_samples_1}
\end{figure}

\begin{figure}[!htbp]
        \centering

        \includegraphics[width=0.9\textwidth]{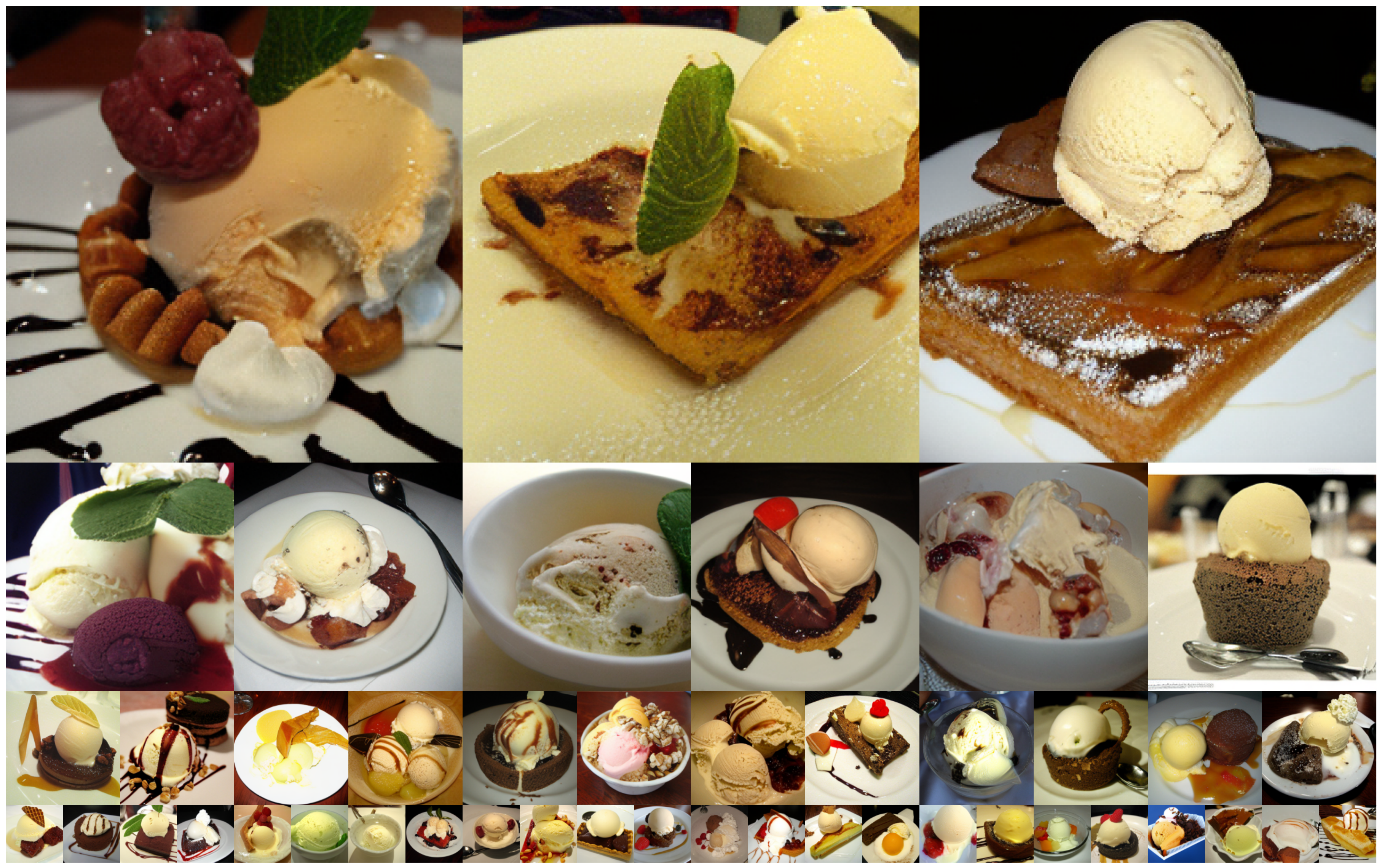} 
    \caption{\textbf{Uncurated generation results} of \texttt{SiT-XL/2} w/ \texttt{ReDi}. We use Classifier-Free Guidance
with $w = 4.0$. Class label = $928$.}
    \label{fig:uncurated_samples_1}
\end{figure}

\begin{figure}[!htbp]
        \centering

        \includegraphics[width=0.9\textwidth]{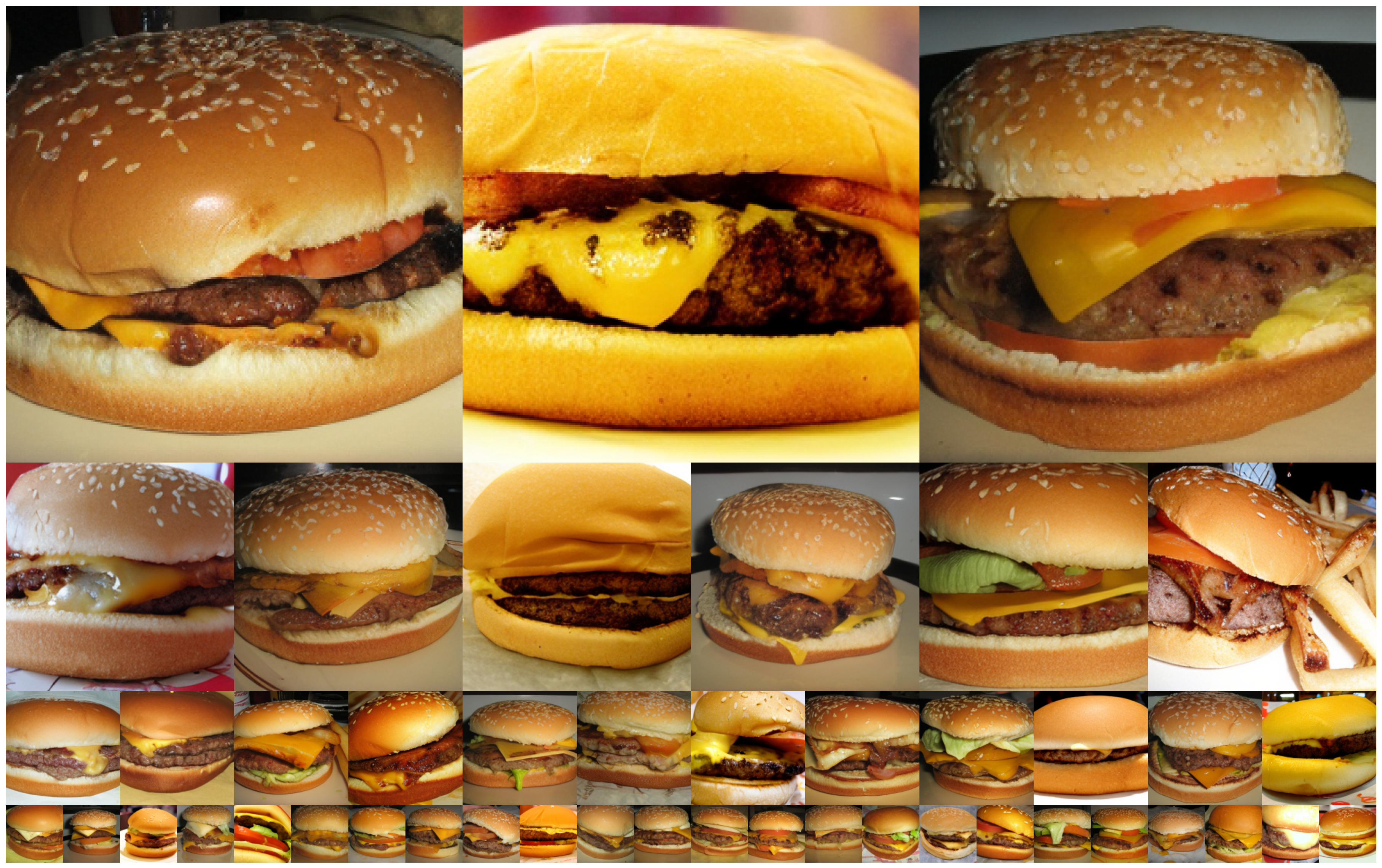} 
    \caption{\textbf{Uncurated generation results} of \texttt{SiT-XL/2} w/ \texttt{ReDi}. We use Classifier-Free Guidance
with $w = 4.0$. Class label = $933$.}
    \label{fig:uncurated_samples_1}
\end{figure}

\end{document}